\pdfoutput=1
\documentclass[11pt]{article}
\usepackage{arabtex}

\usepackage[utf8]{inputenc}
\usepackage{utf8} % Needed by arabtex package

\usepackage{times}
\usepackage{color,soul} 
\usepackage{booktabs}
\usepackage{latexsym}
\usepackage{hyperref}
\usepackage{multirow}
\usepackage[LAE,T1]{fontenc}

\usepackage[utf8]{inputenc} % Needed by LAE, T1 for inline arabic
\DeclareUnicodeCharacter{0424}{\textcyr{F}}
\usepackage[LAE,T1]{fontenc}
\usepackage[arabic,french,english]{babel}
\usepackage[raggedrightboxes]{ragged2e}

\usepackage{todonotes}
\usepackage{lscape}
\usepackage{natbib}
\usepackage{longtable}
\usepackage{enumitem}
\usepackage{float}
\usepackage{makecell}
\usepackage{xcolor, colortbl}
\usepackage{inconsolata}
\usepackage{arydshln}
\usepackage{soul}

\usepackage{graphicx}
\usepackage{scrtime}
\usepackage[dont-mess-around]{fnpct}

\usepackage[ arabic , english]{babel}
\usepackage{colortbl}
\usepackage{etoolbox}
\usepackage{dirtytalk}
\usepackage{array}
\usepackage{pmboxdraw}
\usepackage{bbding}
\usepackage{changepage}
\usepackage{natbib}
\usepackage{times}
\usepackage{tabularx}
\usepackage{latexsym}
\usepackage[utf8]{inputenc}
\usepackage{xcolor}
\usepackage{tcolorbox}
\usepackage[utf8]{inputenc}
\usepackage{graphicx}
\usepackage{comment}
\usepackage{lscape}
\usepackage[T1]{fontenc}
\usepackage[locale=FR]{siunitx}
\usepackage{booktabs}

\usepackage{textcomp}
\usepackage{appendix}

\usepackage{array}
\usepackage{geometry}
\usepackage{multirow}
 \usepackage{array,multirow,graphicx}
 \usepackage{float}
\definecolor{olivegreen}{RGB}{107,142,35}
\definecolor{lightolivegreen}{RGB}{157,192,105}
\definecolor{customgreen}{rgb}{0.13, 0.55, 0.13}

\usepackage{amssymb}% http://ctan.org/pkg/amssymb
\usepackage{pifont}% http://ctan.org/pkg/pifont
\definecolor{blue}{RGB}{0,0,255}
\definecolor{lightblue}{RGB}{0,216,230}
\definecolor{purple}{RGB}{153, 50, 204} % Purple Orchid color

\definecolor{blue}{RGB}{0,0,255}
\definecolor{lightblue}{RGB}{0,216,230}
\usepackage[final]{acl}

\usepackage{times}
\usepackage{latexsym}

\usepackage[T1]{fontenc}
\usepackage[utf8]{inputenc}

\usepackage{microtype}

\usepackage{inconsolata}

\usepackage{graphicx}

\title{
\raisebox{-1.5ex}{\protect\includegraphics[height=4.1\fontcharht\font`\B]{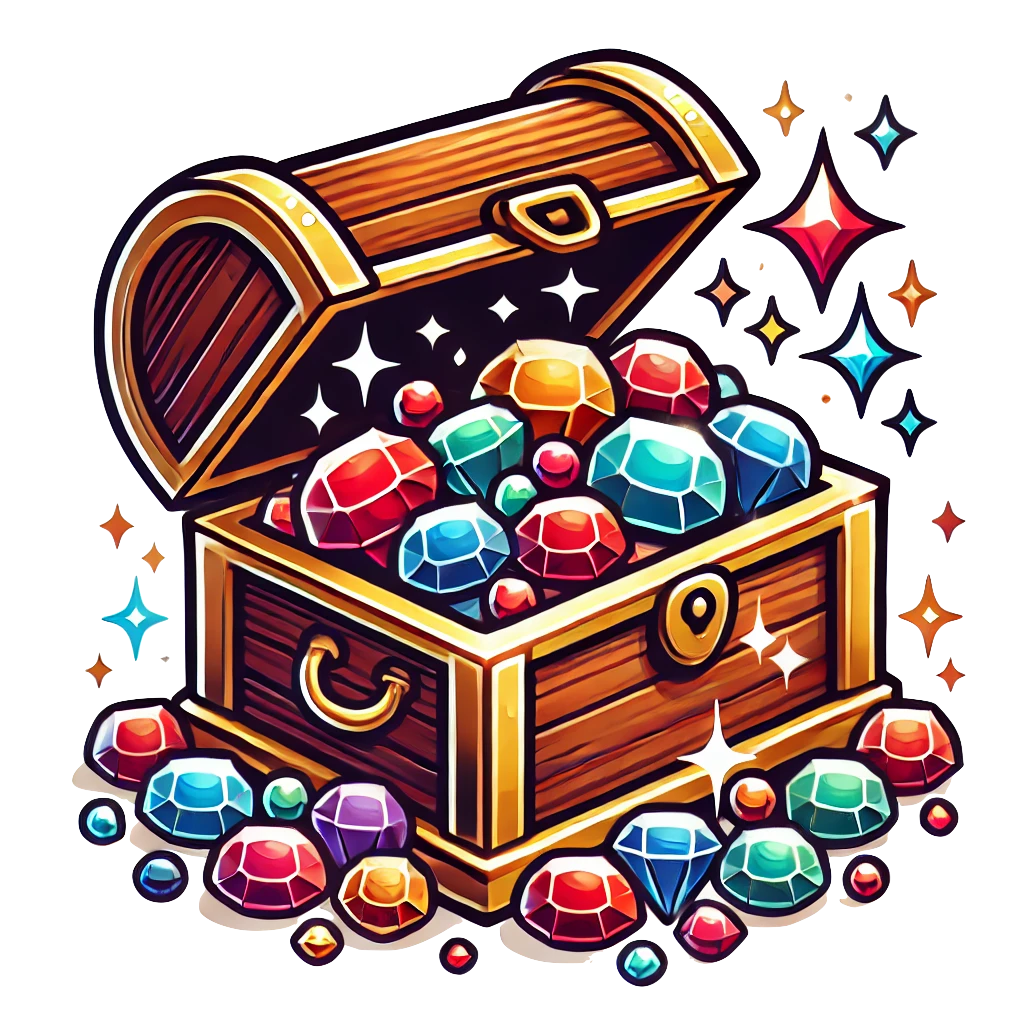}}
\textbf{\textit{\scalebox{1.5}{J}awaher}}:\\[0.5ex]
A Multidialectal Dataset of Arabic Proverbs for LLM Benchmarking
}

\author{\normalsize Samar M. Magdy$^{\xi}$\thanks{~~Equal contribution}~~~~~ Sang Yun Kwon$^{\lambda}$\footnotemark[1]~~~~~~ Fakhraddin Alwajih$^{\lambda}$ \\ 
~\normalsize\textbf{Safaa Abdelfadil}$^{\xi}$ ~\normalsize\textbf{~~~~~~~~Shady Shehata}$^{\xi,\gamma}$ ~\normalsize\textbf{~~~~~~~~Muhammad Abdul-Mageed}$^{\lambda,\xi,\gamma}$~  
 \\
\normalsize $^{\lambda}$The University of British Columbia~~~~ $^{\xi}$MBZUAI~~~~
\normalsize  $^{\gamma}$Invertible AI ~~~~\\
   \texttt{\normalsize {samar.magdy}@mbzuai.ac.ae, {\{skwon01, muhammad.mageed\}}@ubc.ca}
}

\setarab
\setcode{utf8}

\begin{document}
\maketitle
\begin{abstract}
Recent advancements in instruction fine-tuning, alignment methods such as reinforcement learning from human feedback (RLHF), and optimization techniques like direct preference optimization (DPO), have significantly enhanced the adaptability of large language models (LLMs) to user preferences. However, despite these innovations, many LLMs continue to exhibit biases toward Western, Anglo-centric, or American cultures, with performance on English data consistently surpassing that of other languages. This reveals a persistent cultural gap in LLMs, which complicates their ability to accurately process culturally rich and diverse figurative language, such as proverbs. To address this, we introduce \textit{Jawaher}, a benchmark designed to assess LLMs' capacity to comprehend and interpret Arabic proverbs. \textit{Jawaher} includes proverbs from various Arabic dialects, along with idiomatic translations and explanations. Through extensive evaluations of both open- and closed-source models, we find that while LLMs can generate idiomatically accurate translations, they struggle with producing culturally nuanced and contextually relevant explanations. These findings highlight the need for ongoing model refinement and dataset expansion to bridge the cultural gap in figurative language processing. Project GitHub page is accessible at: \href{https://github.com/UBC-NLP/jawaher}{https://github.com/UBC-NLP/jawaher}.
\end{abstract}

\section{Introduction}
Instruction fine-tuning~\cite{chung2024scaling} has significantly enhanced the creativity and customizability of LLMs, while alignment techniques, such as RLHF~\cite{ouyang2022training} and DPO~\cite{rafailov2024direct}, have improved the ability of these models to align with user preferences. The vast reservoir of cultural knowledge embedded within LLMs, combined with the potential of these alignment techniques, has sparked interest in how LLMs can reflect specific human values and personas across different cultures~\cite{gupta2023investigating,kovavc2023large}.

\begin{figure}[t]
  \centering
  \includegraphics[width=\linewidth]{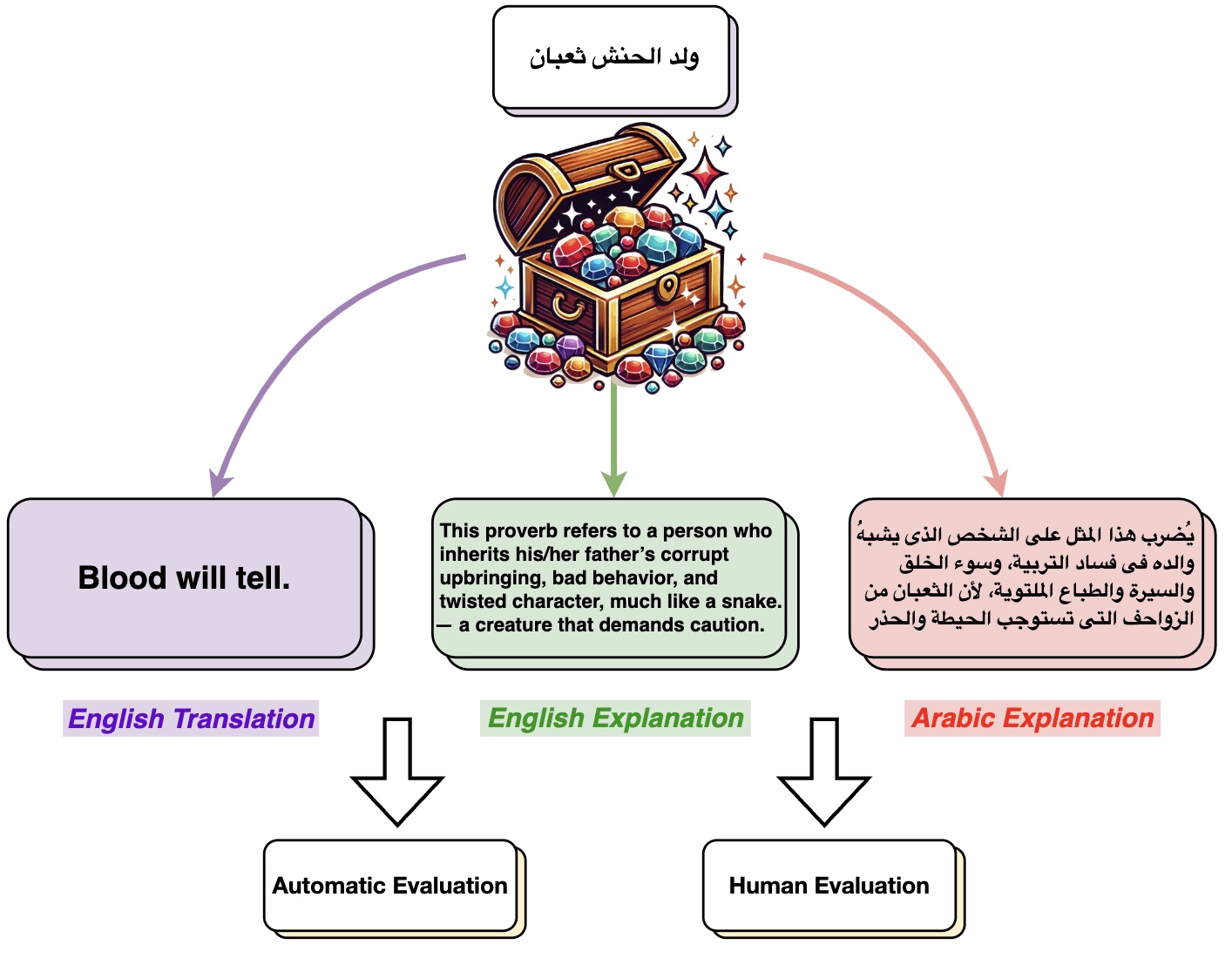}
  \caption{Overview of the \textit{Jawaher} benchmark, featuring Arabic proverbs along with their idiomatic translations and explanations in both English and Arabic; for instance, the \textit{Libyan} proverb \AR{\small{ولد الحنش ثعبان}} whose  literal meaning is "the son of a snake is a serpent", is explained as conveying the idea that "like father, like son in terms of bad behavior and twisted character".}
  \label{fig1}
\end{figure}

Figurative (i.e., non-literal) language~\cite{Fussell,Shutova2011ComputationalAT, gibbs2012interpreting} rich with implicit cultural references, varies significantly across cultures and relies on shared cultural knowledge. Differences in figurative expressions between languages often arise from unique cultural values, historical context, and other regional factors. Thus, understanding figurative language depends on grasping culturally significant concepts and their associated sentiments~\cite{kabra-etal-2023-multi}. To ensure LLMs are inclusive and deployable across regions and applications, it is essential for these models to function adequately in diverse cultural contexts~\cite{adilazuarda2024towards}. Among common types of figurative language, \textit{proverbs} are a key form of cultural expression, encapsulating diverse areas of human knowledge and experience~\cite{gonzalez2002isabel,mieder2004proverbs}. Despite their widespread use, proverbs are linguistically complex due to their cultural significance and structure, exhibiting notable lexical and syntactic variations~\cite{chacoto1994estudo}, making their automatic identification in text challenging for NLP systems.

Recent research aiming to study cultural representation and inclusion in LLMs has found that many models remain strongly biased towards Western, Anglo-centric, or American cultures, with performance on English data still surpassing that of other languages~\cite{dwivedi-etal-2023-eticor,adilazuarda2024towards}. This highlights the existence of significant `\textit{culture gaps}'~\cite{liu2023multilingual} in LLMs, which further complicate their ability to accurately handle culturally rich and diverse forms like proverbs.
In this paper, we make several contributions toward closing the cultural gap in LLMs by introducing \textit{Jawaher}, which facilitates the building and assessment of models designed to understand figurative language, specifically Arabic proverbs, as follows:

 \textbf{\textit{(i) Jawaher}}: a new benchmark for figurative language understanding of Arabic proverbs. \textit{Jawaher} comprises 10,037 high-quality Arabic proverbs with multidialectal coverage across 20 varieties, offering a rich and diverse collection representative of a wide range of Arab countries (Section~\ref{dataset}). The proverbs are paired with idiomatic translations and explanations in English  (Section~\ref{task_representation}) that cover different themes and cultural context (Section~\ref{cultural_context}). Figure~\ref{fig1} illustrates the pairing of proverbs in \textit{Jawaher} with their translation and explanations.
    
\textbf{(ii) Comprehensive experiments to assess \textit{Jawaher}'s usefulness towards building models that understand Arabic proverbs.} We conduct a comprehensive set of experiments using both \textit{open}- and \textit{closed}- source models to test their abilities in interpreting, contextualizing, and translating these proverbs (Section~\ref{model_setup}). We propose extensive automatic and human evaluation to assess model understanding across our proposed tasks (Section~\ref{evaluation_setup}). We find that while models generally perform well on \textit{translation} tasks—producing idiomatically correct outputs—they struggle significantly with \textit{explanation} tasks, particularly in capturing the cultural nuances, historical context, and deeper figurative meanings behind Arabic proverbs. Although closed-source models outperform open-source models, both still face notable limitations in clarity, cultural relevance, and detail in explanations, revealing a substantial gap in fully understanding and conveying the richness of Arabic proverbs (Section~\ref{results_discussion}).

\section{Related Works}
\noindent\textbf{Figurative Language.}
Prior work on figurative language understanding has covered a wide range of topics, including simile detection and generation~\cite{niculae2014brighter, mpouli2017annotating, zeng2020neural}. Results show that, although language models can often recognize non-literal language and may attend to it less, they still struggle to fully capture the implied meaning of such phrases~\cite{shwartz-dagan-2019-still}. This has led to more recent studies shifting toward tasks focused on \textit{comprehending} figurative language~\cite{chakrabarty-etal-2022-flute, he-etal-2022-pre, prystawski2022psychologically, jang2023figurative}. Efforts in dataset and task development have focused on diverse tasks such as recognizing textual entailment, multilingual understanding, and figurative language beyond text~\cite{liu2023multilingual, yosef-etal-2023-irfl,saakyan2024v}. However, the lack of comprehensive datasets focusing on figurative language, particularly Arabic proverbs, limits the ability to thoroughly evaluate LLMs' cultural awareness and their true understanding of culturally embedded non-literal expressions, further motivating our work. More details regarding dataset and task development on figurative language are provided in Appendix~\ref{apendx:development_figlang}. \\

\noindent\textbf{Arabic Proverbs.}
Proverbs are essential repositories of cultural values and historical experiences, conveying general truths or advice using non-literal language~\cite{kuhareva2008arabic,brosh2013proverbs,mieder2021innovative}. Despite their apparent simplicity, proverbs are complex linguistic and cultural products~\cite{mieder2007proverbs}, marked by distinct features that set them apart from ordinary language~\cite{mieder2004proverbs, meliboevich2022influence}. Arabic uniquely maintains a well-defined link between past and present in its linguistic and cultural traditions~\cite{versteegh2014arabic}. Arabic proverbs, rooted in diverse dialects, reflect this continuity, showcasing not only the linguistic richness of the language but also the cultural, historical, and social values of different communities~\cite{karoui2015towards, elmitwally2020classification}. More details on the cultural significance and linguistic complexity of Arabic proverbs are available in the Appendix~\ref{apendx:linguistics_background}. 
\\

\noindent\textbf{Cultural LLM.} LLMs have attracted substantial interest in sociocultural studies, particularly regarding their performance across diverse cultural contexts~\cite{gupta2023investigating,kovavc2023large}. Research has increasingly uncovered a cultural gap, demonstrating that many models are biased toward Western, Anglo-centric perspectives~\cite{johnson2022ghost, liu2023multilingual}. These biases impact linguistic-cultural interactions and challenge value-based objectives~\cite{johnson2022ghost,durmus2023towards}. Efforts to address this include multilingual QA~\cite{kabra-etal-2023-multi}, cross-cultural translation~\cite{singh2024translating}, and culturally diverse dataset creation~\cite{ji2024emma,qian2024cameleval}, all aiming to improve multilingual adaptation and cultural alignment in LLMs.

\section{Jawaher}\label{dataset}
%\fakhr{I created this outline and draft. @Samar please merge this section with with section 6}
\textit{Jawaher} consists of Arabic proverbs paired with their idiomatic or literal English translations, along with explanations in both Arabic and English, covering 20 different Arabic varieties. Below, we outline the analysis of \textit{Jawaher}, focusing on its coverage of dialects, themes, cultural context, and the tasks it facilitates. %Table~\ref{data_stats_proverbs} shows the statistics of \textit{Jawaher}.  

\subsection{\textit{Jawaher} Analysis}

\begin{figure}[t]
  \centering
  \includegraphics[width=\linewidth]{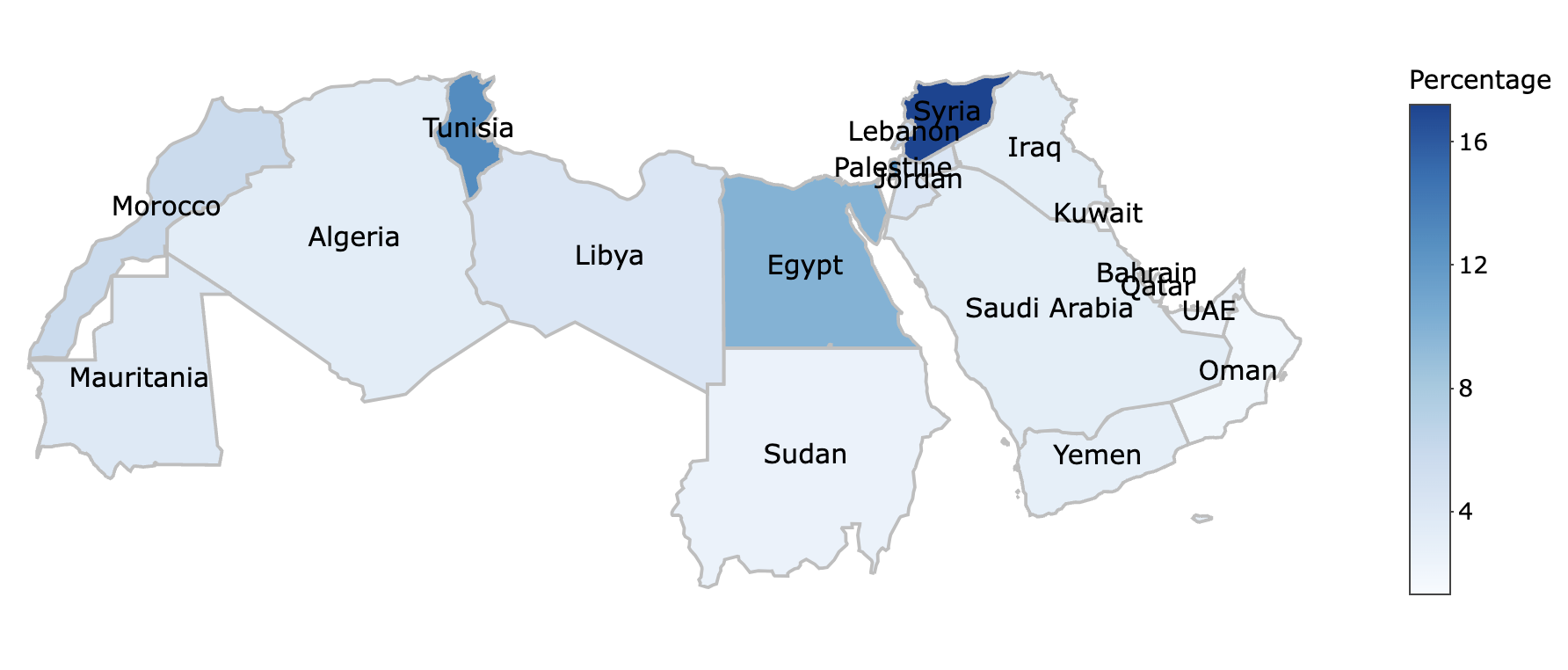}
  \caption{Choropleth map showing the geographical distribution of Arabic varieties covered in  \textit{Jawaher}. Color intensity represents the percentage of proverbs collected from each region, with darker shades indicating higher concentrations.}
  \label{choropleth_map} 
\end{figure}

\noindent\textbf{Dialect Representation.} \label{dialect_representation}
%We collect proverbs from reputable sources to ensure the authenticity and diversity of each dialect. 
Data in \textit{Jawaher} is manually curated by four native Arabic speakers with strong linguistic expertise. The four annotators come from Egypt (two annotators), Mauritania, and Morocco. During data collection, they consulted with native speakers from other countries such as Jordan, Syria, and the United Arab Emirates (UAE) to ensure a diverse and authentic representation of proverbs from countries across the Arab world. We acquire data from publicly available online resources and carefully verify the origins of proverbs to confirm that they truly reflect their respective cultural heritage. Figure~\ref{choropleth_map} shows the geographical distribution of the countries covered in~\textit{Jawaher}, highlighting the dataset’s regional diversity. Modern Standard Arabic (MSA), a pan-Arabic variety not tied to any specific region, is excluded to better showcase the thematic distribution of proverbs across distinct Arab dialects and regions. Below is a list of all covered countries, categorized by dialect group and region: \textbf{\textit{Gulf Dialects:}} Bahraini, Emirati, Kuwaiti, Omani, Qatari, and Iraqi;
\textbf{\textit{Levantine Dialects:}}  Jordanian, Lebanese, Palestinian, and Syrian; 
\textbf{\textit{North African Dialects:}} Algerian, Libyan, Mauritanian, Moroccan, and Tunisian; 
\textbf{\textit{Arabian Peninsula:}} Yemeni and Saudi; and
\textbf{\textit{Nile Basin:}} Egyptian and Sudanese.
%sources and categories
%\begin{itemize}
    %\item \textbf{Literary Works and Anthologies:} Classic and contemporary collections of proverbs specific to each region.
    %\item \textbf{Online Databases:} Trusted websites and digital libraries dedicated to Arabic folklore and idiomatic expressions.
    %\item \textbf{Academic Publications:} Scholarly articles and ethnographic studies focusing on regional sayings and their usages.
    %\item \textbf{Community Contributions:} Input from native speakers via surveys and social media platforms to include less-documented proverbs.
%\end{itemize}

% \begin{itemize} 
%     \item \textbf{Gulf Dialects:} Bahraini, Emirati, Kuwaiti, Omani, Qatari, and Iraqi \item \textbf{Levantine Dialects:}  Jordanian, Lebanese, Palestinian, and Syrian \item \textbf{North African Dialects:} Algerian, Libyan, Mauritanian, Moroccan, and Tunisian \item \textbf{Arabian Peninsula:} Yemeni and Saudi
%     \item \textbf{Nile Basin:} Egyptian and Sudanese
%     \end{itemize}

%The final \textit{Jawaher} dataset encompasses 20 Arabic varieties, ensuring a broad representation of Arabic speaking countries as seen in []:
%\begin{itemize}
%    \item \textbf{North African Dialects:} \hl{Moroccan, Algerian, Tunisian, Libyan, Egyptian, Mauritanian}.
%    \item \textbf{Levantine Dialects:} Lebanese, Syrian, Jordanian, Palestinian.
%    \item \textbf{Gulf Dialects:} Saudi, Emirati, Kuwaiti, Qatari, Bahraini, Omani.
%    \item \textbf{Others:}  Iraqi, Yemeni, MSA.
%\end{itemize} 

\begin{figure*}[]
    \centering
    \includegraphics[width=0.9\textwidth]{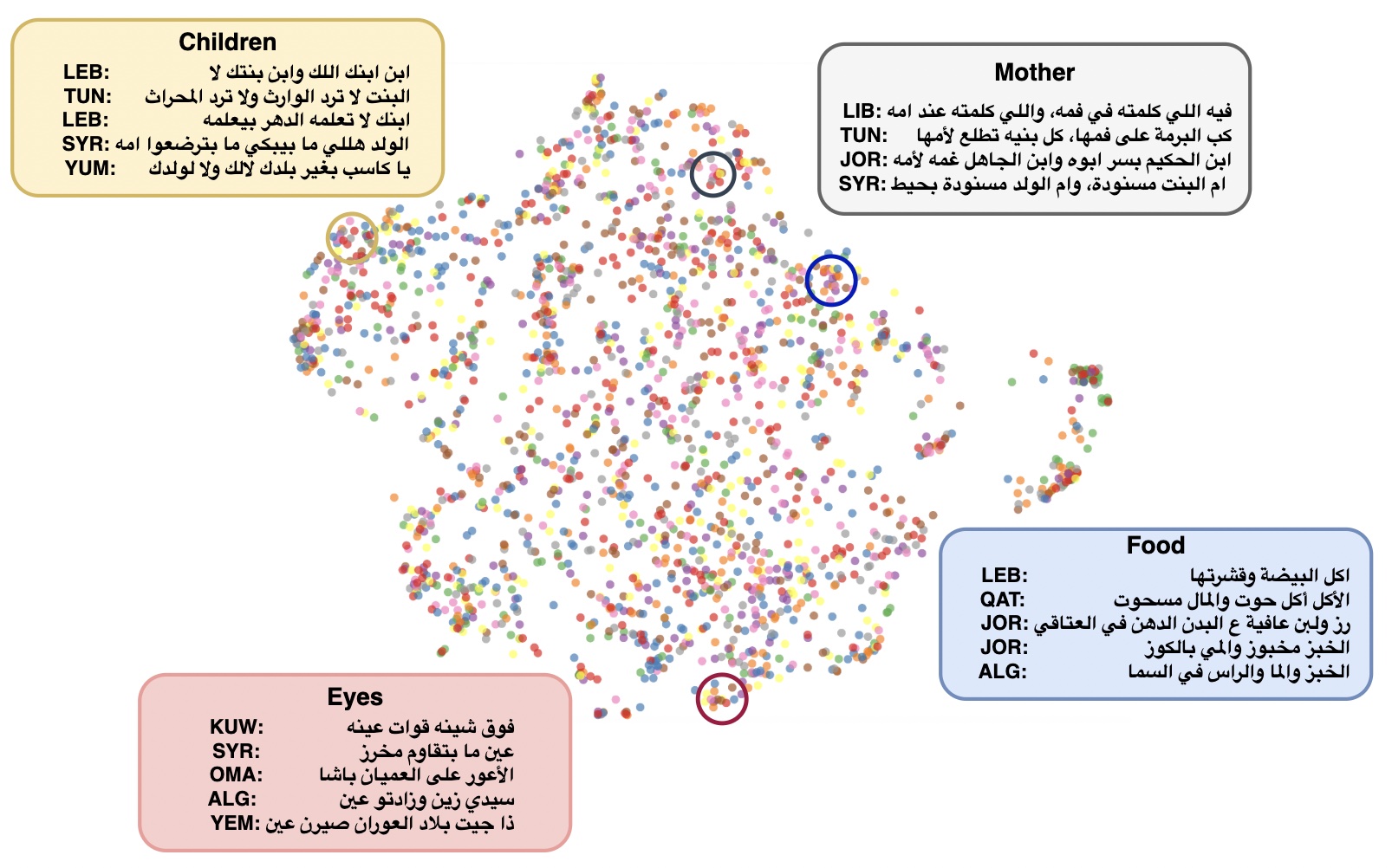}
    \caption{UMAP visualization of proverb embeddings, showing clusters corresponding to four key themes: \textit{Food}, \textit{Mother}, \textit{Children}, and \textit{Eyes}. Proverbs are grouped  based on thematic and cultural similarities across various Arab dialects.} 
    \label{umap_vizualization}
\end{figure*}

\noindent\textbf{Theme Classification.} \label{theme_classification} %Categorization into one or more of the 3,030 identified themes. 
Popular proverbs in Arab countries address various aspects of daily life, serving as an important source of folk culture while reflecting the historical, religious, and societal values of their users. These proverbs often reveal insights into the evolution of civilization and sometimes reference specific events or beliefs related to human existence. Dominant themes frequently incorporate elements such as animals, food, and other culturally significant symbols, to reflect the unique nature and heritage of Arab societies~\cite{farghal2021animal}.

We analyze the themes present in \textit{Jawaher} by collecting embeddings of each proverb using a multilingual text embedding model\footnote{\url{https://huggingface.co/intfloat/multilingual-e5-large}}~\cite{wang2024multilingual}. and perform UMAP visualization on the collected data. Figure~\ref{umap_vizualization} reveals groupings representing different themes, such as food, mother, children, and eye. Upon inspection, these groupings not only bring sentences together by thematic content but also highlight cultural traits specific to each language or variety. \textit{`Body parts'} emerges as a recurring theme, as seen in proverbs from Palestinian, Syrian, and Kuwaiti dialects, where references to `\textit{eyes}' convey both literal meanings and deeper cultural values. The Omani proverb \AR{\small{ الأعور على العميان باشا}} (\textit{Among the blind, the one-eyed is a king}) emphasizes how a person who has one flaw thinks that he is better than someone who has many flaws, as he is better than them. In the Syrian proverb \AR{\small{عين ما بتقاوم مخرز}} (\textit{An eye cannot resist an awl}), we see a metaphor for vulnerability, underscoring the inevitability of hardship. Meanwhile, the Kuwaiti proverb \AR{\small{فوق شينه قوات عينه}} (\textit{On top of his ugliness, he is brazen}) reflects disdain for arrogance despite flaws, emphasizing cultural disapproval of audacity without merit. These examples illustrate how proverbs portray unique cultural traits specific to each society through distinct linguistic expressions.

\begin{figure*}[h]
\centering
\includegraphics[width=1.0\textwidth]{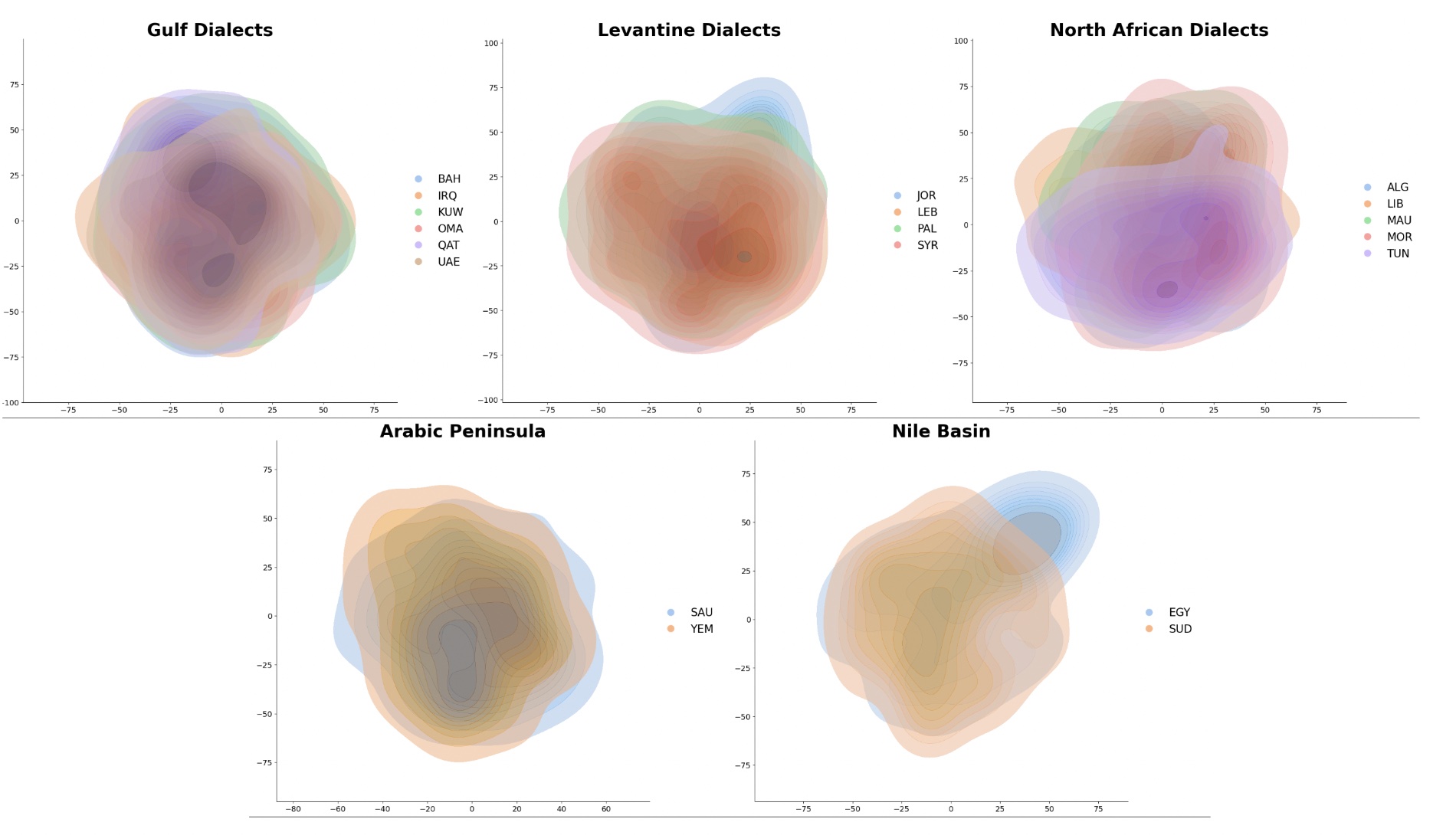}
\caption{ KDE plots of proverb embeddings from various Arabic dialects, highlighting both shared cultural themes and regional variations among  Gulf, Levantine, North African, Arabian Peninsula and Nile Basin dialects.}
\label{group_1}
\end{figure*}

\noindent\textbf{Cultural Context}. \label{cultural_context}
Culture is an integral aspect of language and is crucial for understanding people's backgrounds and social interactions. Without this context, the nuanced messages and symbolic references in proverbs may be lost or misinterpreted~\cite{webster2000arabic}. In structure mapping theory~\cite{gentner1983structure}, figurative language involves comparing a source and target concept, and Arab proverbs often use this mechanism to draw on shared cultural experiences across Arab societies~\cite{kabra-etal-2023-multi, hamdan2023connotations}. The cultural context in Arabic proverbs is thus essential for interpreting their true meanings and appreciating their significance~\cite{bakalla2023arabic}.

As illustrated above, certain themes reflect shared cultural experiences across the Arab world, while the expressions used to convey these themes vary across dialects. To further demonstrate the distinctiveness and connections between proverbs, we plot sentence-level representations using Kernel Density Estimate (KDE) (dimensionality reduction to two components using tSNE~\cite{article-tsne} using sentence embeddings described above. Figure~\ref{group_1} illustrates both shared cultural themes and regional variation across Arabic dialects. Notably, while the central dense areas in all subfigures reflect common cultural values, certain groups show distinct distributions. Proverbs from Gulf dialects, including \textit{Bahraini} (BAH), \textit{Kuwaiti} (KUW), and \textit{Qatari} (QAT), exhibit substantial overlap, reflecting shared cultural and linguistic traditions within the region. Levantine dialects demonstrate strong cohesion, with \textit{Jordanian} (JOR), \textit{Lebanese} (LEB), and \textit{Palestinian} (PAL) proverbs clustering closely, indicating deep cultural interconnections. North African dialects display greater dispersion, particularly among \textit{Algerian} (ALG), \textit{Mauritanian} (MAU), and \textit{Tunisian} (TUN) proverbs, suggesting linguistic diversity within this region.  The Arabic Peninsula  and Nile Basin groups, represented by \textit{Saudi} (SAU) and \textit{Yemeni} (YEM) dialects, and \textit{Egyptian} (EGY) and \textit{Sudanese} (SUD) dialects, respectively, exhibit unique patterns that may reflect historical and sociolinguistic factors.

\subsection{Task Representation} \label{task_representation} Each proverb in \textit{Jawaher} is represented with \textbf{\textit{(a)}} an idiomatic, literal English translation, or English meaning, \textbf{\textit{(b)}} an explanation in Arabic, and \textbf{\textit{(c)}} an explanation in English. The explanations include stories detailing the themes and contexts in which the proverb is commonly used. Explanations, as such, enhance understanding and provide deeper insights into the meanings and cultural significance of proverbs.

To ensure high data quality, entries underwent dual annotation, consensus resolution of discrepancies, and pilot testing to identify and correct any     annotation issues. More details regarding the quality assurance of data are provided in Appendix~\ref{apendx:quality_assurance}.

%\begin{itemize}

\noindent \textbf{Translation of Proverbs.} 
    In \textit{Jawaher}, we provide translations of a total of 4,334 Arabic proverbs, offering different types of translations. All proverbs and their explanations were manually translated by a professional translator to ensure high quality and accurate representation of their meanings. Subsequently, two additional professional translators reviewed and revised the translations to verify their correctness and ensure they were correctly expressed idiomatically. We provide three types of translations of the proverbs: First, we provide (1) \textit{English equivalents} for $1,289$ Arabic proverbs—these are similar idiomatic and figurative expressions in English. Since proverbs are inherently idiomatic, their translations should use common phrases and expressions of the target language \cite{baker2018other,gorjian2024translating}, even if that means diverging from a literal translation to better capture the original's figurative or stylistic essence. Then, we offer (2) $840$ \textit{English meanings} of proverbs by paraphrasing them when \textit{English equivalent} translations were not available. Finally, we provide (3) $2,205$ literal translations. Idiomatic expressions like proverbs are notoriously difficult to translate accurately \cite{gorjian2024translating}. When direct equivalents in the target language are unavailable, translators resorted to providing literal translations. We believe that the inclusion of literal translations is essential in our work as it allows understanding the original wording and cultural nuances of the proverbs, which may not be fully captured through idiomatic or equivalent translations alone.

\noindent \textbf{Arabic Explanation.}
    Our dataset includes $3,764$ Arabic explanations that cover the stories behind the proverbs, how the proverbs are used in certain dialects, and the different situations in which these proverbs may be employed. Furthermore, for certain dialects such as \textit{Lebanese}, and \textit{Omani}, explanations include meanings of unusual words to provide more information for the reader to understand what these words mean in MSA.

\noindent \textbf{English Explanation.} \textit{Jawaher} includes $2,500$ human-translated English explanations of Arabic proverbs. These explanations cover proverbs with historical backgrounds or cultural stories from the Arab world, and we aim to have these explanations accurately convey the meanings behind the proverbs.

Full statistics for each task and examples across all dialects are provided in the Appendix~\ref{apendx:data_statistics_full}.

\section{Experimental Setup}
\begin{figure}[t]
  \centering
  \includegraphics[width=\linewidth]{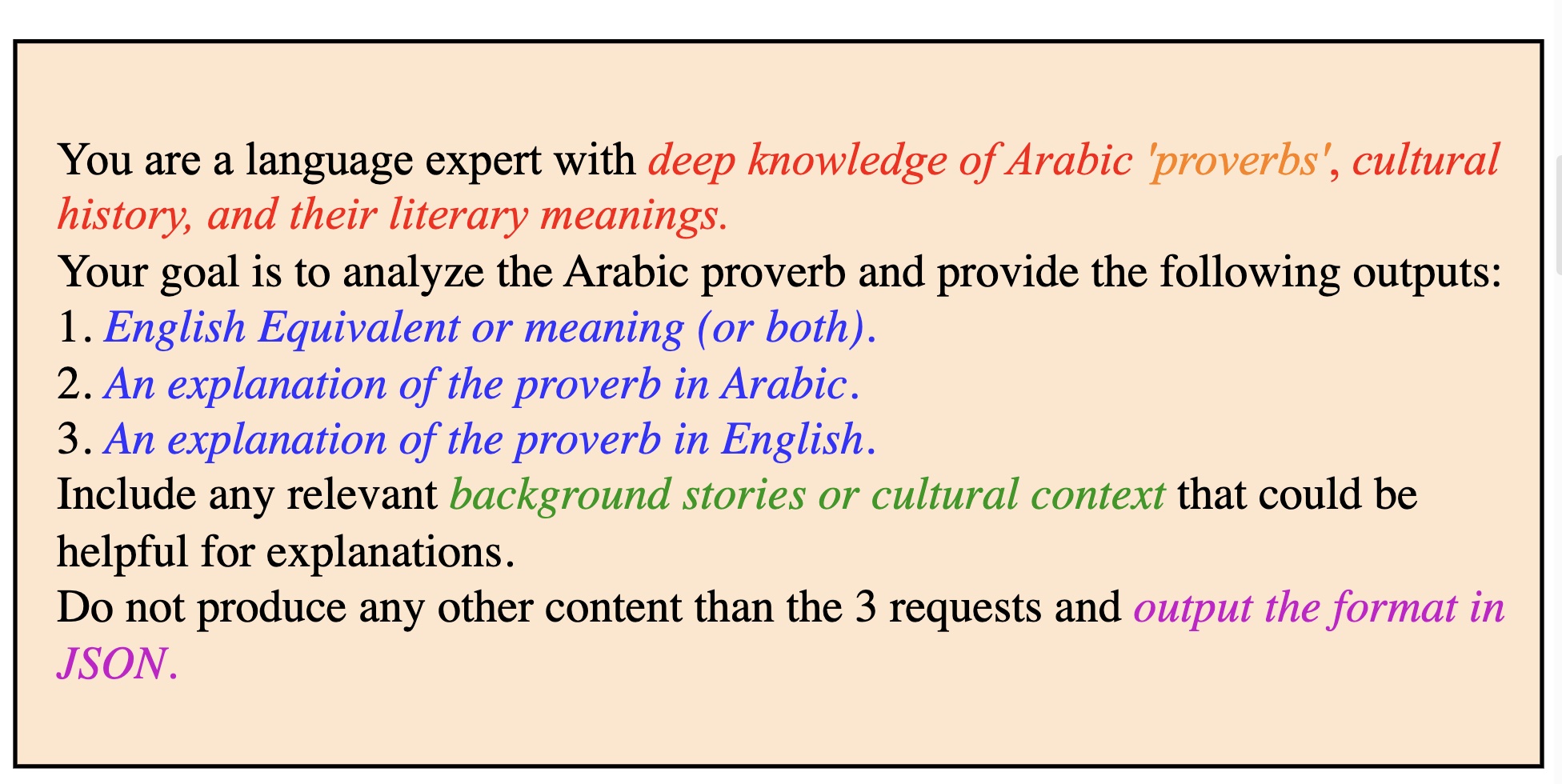}     
  \caption{Designed prompt used to test model's performance on \textit{Jawaher}.}
  \label{prompt} 
\end{figure}

We evaluate \textit{Jawaher} using both \textit{open}- and \textit{closed}-source state-of-the-art multilingual LLMs (mLLMs) to assess their abilities across our proposed tasks. The models are tested in a zero-shot setting~\cite{sanh2021multitask}, allowing us to evaluate their inherent capacity to interpret, explain, and contextualize Arabic proverbs. To achieve this, we create a universal prompt template in English: (1) We set the {\color{red}\textit{role}} of the model as a language expert with deep knowledge of Arabic proverbs, cultural history, and literary meanings. (2) We deliver the {\color{orange}\textit{test input}} to the model to produce the output. (3) We provide the name of the {\color{blue}\textit{task}} that needs to be performed. (4) We specify the {\color{customgreen}\textit{context}}, asking to include any relevant background stories or cultural context that could be helpful for explanations on the tasks. (5) Finally,  we define what should be the expected {\color{purple}\textit{outcome}} of the model. The prompt used to evaluate mLLMs can be found in Figure~\ref{prompt}.

\subsection{Models}\label{model_setup}
\noindent \textbf{Open Source.} For \textit{open}- source mLLMs, we evaluate two models from the Llama 3 family~\cite{dubey2024llama}: \texttt{Llama-$3.1$-$8$B-Instruct}, \texttt{Llama-$3.2$-$3$B-Instruct}, and \texttt{Google's Gemma-$2$-$9$B-it}~\cite{team2024gemma}. These models were chosen for their varying sizes and multilingual capabilities, including support for Arabic.

\noindent \textbf{Closed Source.} We experiment with the following \textit{closed}- source leading mLLMs: \texttt{GPT-4o}~\cite{achiam2023gpt}, \texttt{Gemini 1.5 Pro}~\cite{reid2024gemini}, \texttt{Claude 3.5 Sonnet}\footnote{\href{https://www.anthropic.com/news/claude-3-5-sonnet}{https://www.anthropic.com/claude-3-5-sonnet}}, and \texttt{Cohere Command R+}\footnote{\url{https://docs.cohere.com/docs/command-r-plus}}.

\subsection{Evaluation setup} \label{evaluation_setup}
To evaluate the aforementioned models, we sample ten examples from each dialect and prompt the model using our crafted prompt. Below, we describe the automatic metrics and criteria we employ for human evaluation in order to assess the model's capabilities to understand and interpret the figurative language of Arabic proverbs across multiple dialects and cultural contexts.

\noindent\textbf{Automatic Evaluation.} We use BLEURT~\cite{sellam-etal-2020-bleurt} and BERTScore~\cite{zhang2019bertscore} to judge the quality of explanations and translations. These metrics compute token similarity using contextual embeddings instead of exact matches, making them appropriate for measuring \textit{fluency} and conveying the reference's \textit{meaning}. Since BLEURT is calculated at the sentence level, we average these scores over the entire test set and report the final score on a scale between 0 and 100. BLEURT also only supports English; therefore, we only evaluate English-related tasks on BLEURT. For BERTScore, we report only the F\textsubscript{1} score. Full scores can be found in Appendix~\ref{apendx:automatic_evalaution}. 

\noindent\textbf{Human Evaluation.} Along with automatic evaluation, we conduct human evaluation to assess model performance. We develop evaluation criteria tailored to the nature of our tasks. Human evaluation is carried out by two expert annotators, both holding degrees in linguistics and translation studies, to ensure the accuracy and consistency of the evaluation criteria.  For the first task, \textit{translation}, we assess the models based on \textit{accuracy} and \textit{idiomaticity}, as the task inherently involves figurative language. Our goal is to determine how well the models understand the figurative aspects of the proverbs, translate them correctly, and convey their intended meanings. For the second task, providing \textit{explanations} in both Arabic and English, we evaluate model output across four metrics: \textit{clarity}, \textit{depth \& detail}, \textit{correctness}, and \textit{cultural relevance and sensitivity}. Each annotator evaluated the models' performance using the developed metrics, by scoring each response on a scale from 1 to 5. Appendix~\ref{apendx:human_evaluation_metrics} outlines our evaluation metrics in more detail. 
\noindent\textbf{LLM as Judge.} Recently, LLM-based evaluation~\cite{zeng2023evaluating, chern2024can} has emerged as a scalable and cost-effective alternative to human evaluations. In this work, we use \textit{LangChain's}\footnote{\href{https://python.langchain.com/v0.1/docs/guides/productionization/evaluation/}{https://python.langchain.com/v0.1/docs}} evaluation framework with \textit{string evaluator}. We specifically customize the evaluation criteria to our own and use \texttt{GPT-$4$o} for evaluation.

\section{Results \& Discussion} \label{results_discussion}
Table~\ref{automatic_evaluation} reports BERTScores for all three tasks and BLEURT scores for \textit{English translation} and \textit{explanation} tasks.

\begin{table}[]
\centering
\renewcommand{\arraystretch}{1.5}
\resizebox{\columnwidth}{!}{%
\begin{tabular}{llcc|cc|c}\toprule
\multirow{2}{*}{\textbf{Source}} & \multirow{2}{*}{\textbf{Models}} & \multicolumn{2}{c}{\textbf{En. Trans.}} & \multicolumn{2}{c}{\textbf{En. Exp.}} & \textbf{Ar. Exp.} \\\cmidrule{3-4}\cmidrule{5-6}\cmidrule{7-7}
 &  & \textbf{F\textsubscript{$1$}} & \textbf{BLEURT} & \textbf{F\textsubscript{1}} & \textbf{BLEURT} & \textbf{F\textsubscript{1}} \\\toprule
\multirow{3}{*}{\textbf{Open}} 
& Llama $3.1$ & $89.42$~ & $30.12$~ & $89.19$~ & \underline{$46.51$}~ & $67.05$~ \\
& Llama-$3.2$ & $89.48$~ & $27.56$~ & $89.23$~ & $43.16$~ & $66.10$~ \\
& Gemma & \underline{$90.74$}~ & \underline{$33.37$}~ & \underline{$90.64$}~ & $44.50$~ & \underline{$67.22$}~ \\\midrule
\multirow{4}{*}{\textbf{Closed}} 
& GPT-4o & \underline{\textbf{91.22}}~ & $44.50$~ & \underline{\textbf{94.17}}~ & $40.68$~ & \underline{\textbf{69.21}}~ \\
& Gemini & $89.73$~ & $38.53$~ & $89.69$~ & \underline{\textbf{49.77}}~ & $68.62$~ \\
& Comm. R+ & $90.09$~ & $39.91$~ & $89.93$~ & $45.36$~ & $68.34$~ \\
& Claude & $90.64$~ & \underline{\textbf{44.70}}~ & $83.62$~ & $48.95$~ & $69.06$~ \\
\bottomrule
\end{tabular}%
}
\caption{BERTScore, reported as F\textsubscript{1}, and BLEURT scores for english translation and english explanation. \underline{Scores} represent the highest values in each source category, with the highest across all source categories highlighted in \textbf{bold}. Gemma: Gemma-$2$-$9$B-instruct. Gemini: Gemini $1.5$ Pro. Claude: Claude $3.5$ Sonnet.}\label{automatic_evaluation}
\end{table}

\subsection{Automatic Evaluation}
Among open-source models, \texttt{Gemma2-9B-it} consistently outperforms both Llama models across all tasks in both metrics. It achieves the highest BLEURT scores in \textit{English translation} ($33.37$) and \textit{English explanation} ($44.50$) tasks. This aligns with the BERTScore results, where it shows the best performance among open-source models.

For closed-source models, \texttt{Claude $3.5$ Sonnet} demonstrates the highest BLEURT score for \textit{English translation} ($44.70$), while \texttt{Gemini $1.5$ Pro} leads in \textit{English explanation} ($49.77$). This contrasts with the BERTScore results, where \texttt{GPT-$4$o} shows superior performance across all tasks. Notably, \texttt{GPT-$4$o} also remains strong on BLEURT for English Translation ($44.50$), though it is slightly behind \texttt{Claude $3.5$ Sonnet} on that metric.

Looking across tasks, BLEURT scores for Arabic explanation reveal a noticeable performance drop for all models, suggesting that Arabic figurative language poses a greater challenge than English. While closed-source models generally outperform open-source ones, the performance differences vary by metric and task. This variability underscores the limitations of automatic evaluation when gauging how well models handle figurative language across different languages and prompts.

\begin{figure}[t]
  \centering
  \includegraphics[width=\linewidth]{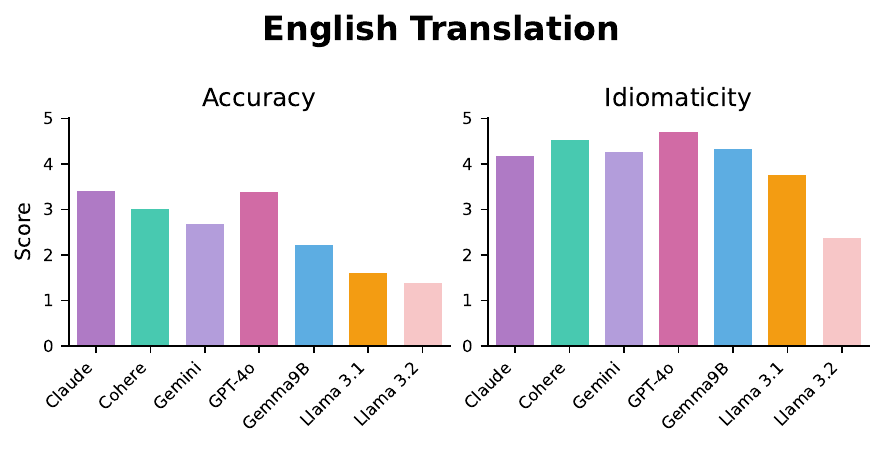}    
  \caption{ Results of human evaluation for both open- and closed-source models on two criteria in the English Translation task.}
  \label{english_equivalent} 
\end{figure}

\begin{figure*}[h]
\centering
\includegraphics[width=0.90\textwidth]{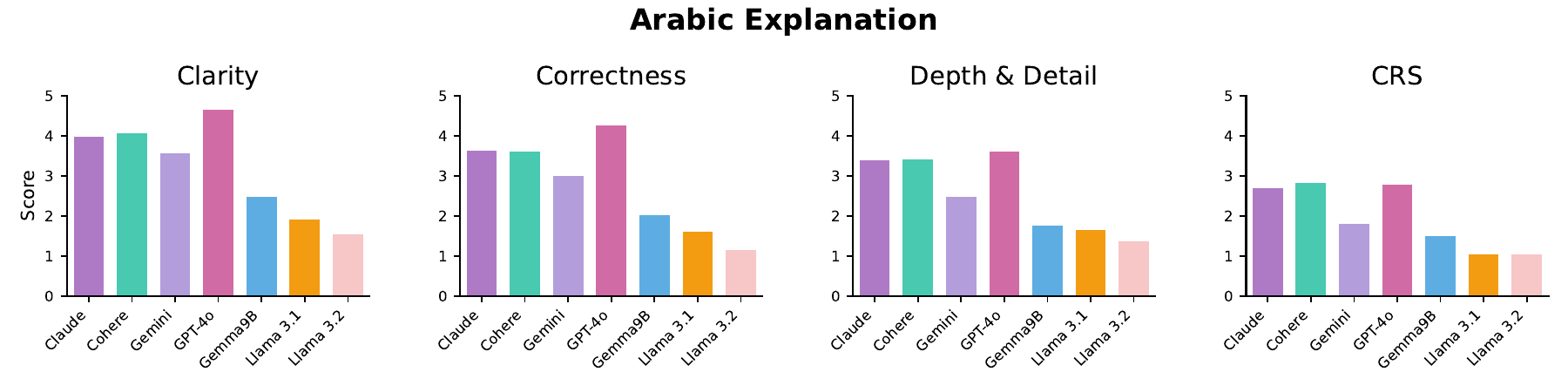}
\caption{ Results of human evaluation for both open- and closed-source models on four criteria in the Arabic explanation task. \textbf{CRS}: cultural relevance and sensitivity.}
\label{arabic_explanation}
\end{figure*}

\begin{figure*}[h]
\centering
\includegraphics[width=0.90\textwidth]{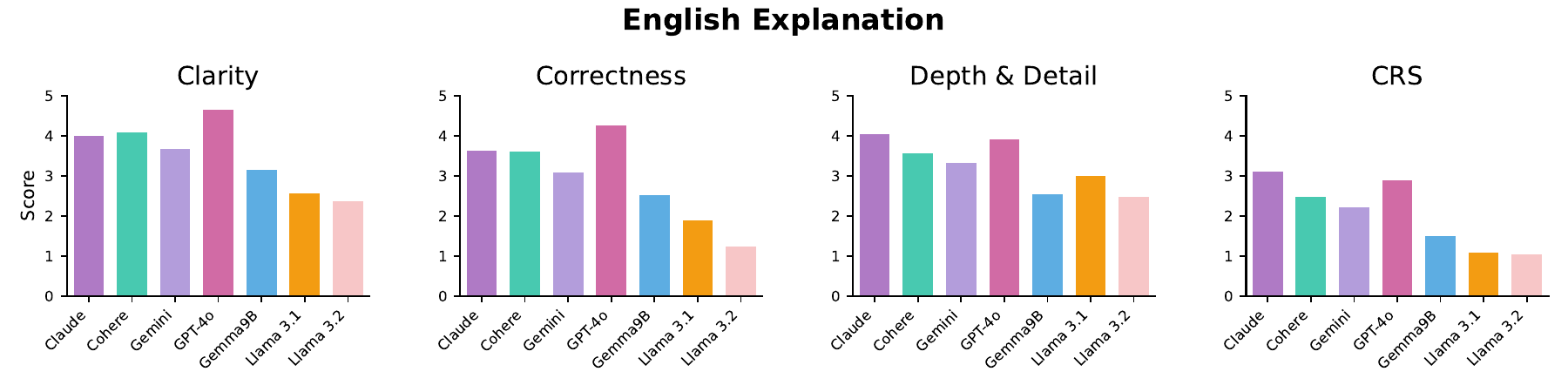}
\caption{Results of human evaluation for both open- and closed-source models on four criteria in the English explanation task. \textbf{CRS}: cultural relevance and sensitivity.}
\label{english_explanation}
\end{figure*}

\subsection{Human Evaluation}
Two annotators were recruited for human evaluation, and we used \textit{LangChain's} evaluation framework to simulate human annotation as described above. IAA using Krippendorff’s $\alpha$~\cite{krippendorff2011alpha}  shows high agreement between the two human annotators, but a significant drop in correlation when including the third annotator (GPT-$4$o). Similarly, there is a notable drop in agreement scores for both the \textit{English and Arabic explanation} tasks, with drops to $0.39$ and $0.20$, respectively. Consequently, we exclude model evaluations and report the average scores between the two annotators. The full IAA score is in Table~\ref{inter_annotator}.

\begin{table}[t]
\centering
\resizebox{\columnwidth}{!}{%
\begin{tabular}{lcc}
\toprule
\textbf{Criteria} & \textbf{Human-only} & \textbf{Human-LLM} \\
\midrule
Accuracy & 0.934 & 0.750 \\
Idiomaticity & 0.957 & 0.195 \\
Clarity (Ar. exp.) & 0.701 & 0.381 \\
Correctness (Ar. exp.) & 0.966 & 0.351 \\
Depth and Detail (Ar. exp.) & 0.976 & 0.547 \\
Cultural rel. \& Sensitivity (Ar. exp.) & 0.703 & 0.050 \\
Clarity (En. exp.) & 0.155 & 0.289 \\
Correctness (En. exp.) & 0.949 & 0.678 \\
Depth and Detail (En. exp.) & 0.400 & 0.536 \\
Cultural rel. \& Sensitivity (En. exp.) & 0.773 & 0.405 \\
\bottomrule
\end{tabular}%
}
\caption{Krippendorff's inter-annotator scores for different criteria across annotator groups for human-only agreements and human-LLM combined agreement.}
\label{inter_annotator}
\end{table}

\noindent\textbf{Translation Quality.} Figure~\ref{english_equivalent} shows \textit{accuracy} and \textit{idiomaticity} scores for the \textit{translation} task. Among closed-source models, \texttt{GPT-$4$o} achieves the highest score of ($3.04$) on \textit{accuracy}. For open-source models, \texttt{Gemma2-9B-it} scores highest with $2.06$. The Llama models perform the worst, with \texttt{Llama $3.1$} and \texttt{Llama $3.2$} scoring ($1.55$) and ($1.36$), respectively. This underperformance highlights a significant limitation in these models when producing nuanced linguistic and cultural content.

\textit{Idiomaticity} scores, highlight models' ability to produce natural and fluent translations. \texttt{GPT-$4$o} scores highest among all models tested with a score of ($4.81$). Other closed-source models also score high, with \texttt{Cohere Command R+}, \texttt{Claude $3.5$ Sonnet}, and \texttt{Gemini $1.5$ Pro} scoring ($4.66$), ($4.59$), and ($4.58$) respectively. Noticeably \texttt{Gemma2-9B-it} scores closely with closed-source models on this criterion, achieving ($4.44$). However, Llama models continue to show lower performance, scoring lowest among all models. Generally, both open- and closed-source models score higher on \textit{idiomaticity}, indicating their ability to produce idiomatically correct translations. Regarding open-source models, we find that they often produce idiomatic expressions that are incorrect and fail to align with the meaning of the proverbs.

\noindent\textbf{Arabic Explanation Quality.}  
Figure~\ref{arabic_explanation} presents the results for the \textit{Arabic explanation} task. \texttt{GPT-$4$o} scores highest in \textit{clarity} ($4.65$), \textit{correctness} ($4.26$), and \textit{depth \& details} ($3.61$), demonstrating its ability to deliver clear, accurate, and comprehensive explanations with underlying meaning and cultural nuances. \texttt{Gemini $1.5$ Pro} scores lowest among closed- models, particularly in \textit{cultural relevance and sensitivity} ($1.80$). Open-source models perform poorly across all criteria, with \texttt{Gemma2-9B-it} achieving the highest score among them at ($2.47$) in \textit{clarity}, while the Llama models consistently score below 2 in all criteria.

Overall, closed-source models, particularly \texttt{GPT-$4$o}, outperformed across all \textit{Arabic explanation} task metrics, excelling in generating clear, correct, and detailed explanations with cultural sensitivity. Although these models scores relatively high on \textit{clarity} and \textit{correctness}, their lower relative scores in \textit{depth \& detail} and \textit{cultural relevance and sensitivity} imply these models are less capable of providing nuanced and context-rich explanations required to fully understand the proverbs. Open-source models showed significantly lower performance in explaining the stories behind the proverbs, their usage in certain dialects, and the situations in which these proverbs are used. Their explanations are generally short and lack cultural background stories. 

\noindent\textbf{English Explanation Quality.}  
Figure~\ref{english_explanation} presents the results for the \textit{English explanation} task. \texttt{GPT-$4$o} continues to score highest in \textit{clarity} ($4.65$) and \textit{correctness} ($4.25$), similar to its results in the \textit{Arabic explanation} task, followed by \texttt{Cohere Command R+} and \texttt{Claude $3.5$ Sonnet}. For \textit{depth \& details}, \texttt{Claude $3.5$ Sonnet} led with ($4.04$), and it also scores highest in \textit{cultural relevance and sensitivity} ($3.11$). Open-source models, including \texttt{Gemma2-9B-it} and the Llama family models, perform significantly lower across all metrics, with \texttt{Gemma2-9B-it} achieving the best score among them in \textit{clarity} ($3.15$).

Overall, closed-source models, particularly \texttt{GPT-$4$o} and \texttt{Claude $3.5$ Sonnet}, outperform across all metrics, while open-source models lag, especially in \textit{cultural relevance and sensitivity} and \textit{depth \& detail} in \textit{explanation} tasks, indicating areas for improvement. English explanations are more detailed, with some reflecting cultural relevance, but none include the stories behind the proverbs. Results from both \textit{English and Arabic explanation} tasks imply that models struggle to provide the nuanced and context-rich explanations required for fully understanding the proverbs in both languages.

\section{Conclusion}
We introduced \textit{Jawaher}, a new benchmark for figurative language understanding in Arabic proverbs. Our manually curated dataset features proverbs from various Arabic dialects, paired with idiomatic translations and explanations, facilitating the development and evaluation of models capable of translating and explaining figurative language. Our experiments show that while both open- and closed-source mLLMs are able to provide idiomatically correct translations, they struggle with the deeper challenge of comprehending figurative language, particularly  when generating detailed explanations that capture the stories behind the proverbs, dialect-specific nuances, and historical or cultural contexts discussed in our dataset. These findings highlight the need for continuous model improvements and dataset enrichment to address the complexities of fully understanding proverbs. We hope our work will inspire further interest in figurative language understanding, particularly across diverse Arabic dialects.

\section{Limitations}
We identify the following limitations in this work:
\begin{itemize}
    \item This study primarily evaluates models under a zero-shot setting, which provides insights into their inherent abilities to understand and process Arabic proverbs without additional training. However, this approach may not reflect their full potential, particularly when it comes to more complex figurative language tasks. Future work could explore fine-tuning these models on proverb-specific datasets to enhance their performance.
    \item The human evaluation process, particularly in the context of cultural nuances, can be subjective and influenced by the evaluators’ individual biases. This is especially relevant when assessing culturally grounded tasks like proverb explanation, where personal interpretation of cultural and regional references may affect the consistency and reliability of judgments.
    \item While the models used in this study claim to include Arabic in their training data, they are not fully optimized for Arabic or its dialectal variations. This limitation may have adversely impacted performance, especially when dealing with proverbs that rely heavily on unique linguistic and cultural nuances found across different Arabic-speaking regions.
\end{itemize}

\section*{Ethics Statement}
\noindent\textbf{Promoting Cultural Sensitivity and Representation in AI.}
Proverbs are deeply embedded in cultural contexts, epitomizing the wisdom, values, and traditions of a community. Recognizing the cultural significance of Arabic proverbs, our research prioritizes the accurate representation and interpretation of these expressions within LLMs. \textit{Jawaher} includes proverbs from various Arabic dialects, ensuring a broad and inclusive representation of the linguistic diversity inherent in the Arabic-speaking world. By addressing Anglo-centric biases prevalent in many LLMs, our work promotes cultural sensitivity and strives to create AI systems that are more equitable and reflective of diverse cultural narratives. This focus on cultural representation is important for developing AI tools that are both effective and respectful of the rich cultural heritage they aim to serve.

\noindent\textbf{Advancing the Understanding of Figurative Language in AI.}
Proverbs represent a form of figurative language that poses unique challenges for automated interpretation and generation by LLMs. Our evaluation of \textit{Jawaher} highlights the current limitations of LLMs in capturing the cultural nuances and contextual relevance inherent in proverbs. 
This focus on figurative language  advances the technical capabilities of AI systems and contributes to a deeper understanding of how language models can better align with human cultural and linguistic diversity.

\noindent\textbf{Data Privacy.}
 \textit{Jawaher} dataset comprises proverbs that are publicly available and do not contain any personal or sensitive information. By utilizing data from public sources, we eliminate privacy concerns and ensure compliance with ethical guidelines and institutional policies regarding data usage. 
\section*{Acknowledgments}\label{sec:acknow}
We acknowledge support from Canada Research Chairs (CRC), the Natural Sciences and Engineering Research Council of Canada (NSERC; RGPIN-2018-04267), the Social Sciences and Humanities Research Council of Canada (SSHRC; 895-2020-1004; 895-2021-1008), Canadian Foundation for Innovation (CFI; 37771), Digital Research Alliance of Canada,\footnote{\href{https://alliancecan.ca}{https://alliancecan.ca}} and UBC Advanced Research Computing-Sockeye.\footnote{\href{https://arc.ubc.ca/ubc-arc-sockeye}{https://arc.ubc.ca/ubc-arc-sockeye}}

\bibliography{custom}

\begin{thebibliography}{71}
\providecommand{\natexlab}[1]{#1}

\bibitem[{Abdul-Mageed et~al.(2020)Abdul-Mageed, Zhang, Bouamor, and Habash}]{abdul-mageed-etal-2020-nadi}
Muhammad Abdul-Mageed, Chiyu Zhang, Houda Bouamor, and Nizar Habash. 2020.
\newblock \href {https://aclanthology.org/2020.wanlp-1.9} {{NADI} 2020: The first nuanced {A}rabic dialect identification shared task}.
\newblock In \emph{Proceedings of the Fifth Arabic Natural Language Processing Workshop}, pages 97--110, Barcelona, Spain (Online). Association for Computational Linguistics.

\bibitem[{Achiam et~al.(2023)Achiam, Adler, Agarwal, Ahmad, Akkaya, Aleman, Almeida, Altenschmidt, Altman, Anadkat et~al.}]{achiam2023gpt}
Josh Achiam, Steven Adler, Sandhini Agarwal, Lama Ahmad, Ilge Akkaya, Florencia~Leoni Aleman, Diogo Almeida, Janko Altenschmidt, Sam Altman, Shyamal Anadkat, et~al. 2023.
\newblock \href {"https://arxiv.org/abs/2303.08774"} {Gpt-4 technical report}.
\newblock \emph{arXiv preprint arXiv:2303.08774}.

\bibitem[{Adilazuarda et~al.(2024)Adilazuarda, Mukherjee, Lavania, Singh, Dwivedi, Aji, O'Neill, Modi, and Choudhury}]{adilazuarda2024towards}
Muhammad~Farid Adilazuarda, Sagnik Mukherjee, Pradhyumna Lavania, Siddhant Singh, Ashutosh Dwivedi, Alham~Fikri Aji, Jacki O'Neill, Ashutosh Modi, and Monojit Choudhury. 2024.
\newblock \href {https://arxiv.org/abs/2403.15412} {Towards measuring and modeling" culture" in llms: A survey}.
\newblock \emph{arXiv preprint arXiv:2403.15412}.

\bibitem[{Ahmed(2005)}]{ahmed2005educational}
Salwa Ahmed. 2005.
\newblock \href {http://harep.org/Social%20Science/ahmed_salw.pdf} {\emph{Educational and social values expressed by proverbs in two cultures: knowledge and use of proverbs in Sudan and in England}}.
\newblock Ph.D. thesis, Berlin, Techn. Univ., Diss., 2005.

\bibitem[{Al~Issawi(2021)}]{al2021cultural}
Juhaina~Maen Al~Issawi. 2021.
\newblock \href {https://www.sciedupress.com/journal/index.php/elr/article/view/20320} {Cultural-bound meaning of animal names in arabic}.
\newblock \emph{English Linguistics Research}, 10(1):42--55.

\bibitem[{Almanea(2021)}]{almanea2021automatic}
Manar~M Almanea. 2021.
\newblock \href {https://ieeexplore.ieee.org/abstract/document/9585619/} {Automatic methods and neural networks in arabic texts diacritization: a comprehensive survey}.
\newblock \emph{IEEE Access}, 9:145012--145032.

\bibitem[{Aransa(2015)}]{aransa2015statistical}
Walid Aransa. 2015.
\newblock \href {https://theses.hal.science/tel-01316544/} {\emph{Statistical machine translation of the Arabic language}}.
\newblock Ph.D. thesis, Universit{\'e} du Maine.

\bibitem[{Bakalla(2023)}]{bakalla2023arabic}
Muhammad~Hasan Bakalla. 2023.
\newblock \href {https://books.google.ae/books?hl=en&lr=&id=kQ7AEAAAQBAJ&oi=fnd&pg=PP1&dq=%7Bbakalla2023arabic,+++title%3D%7BArabic+culture:+through+its+language+and+literature%7D,+++author%3D%7BBakalla,+Muhammad+Hasan%7D,+++year%3D%7B2023%7D,+++publisher%3D%7BTaylor+%5C%26+Francis%7D&ots=rWLnEUVW8I&sig=b6Kx-J_0MaHF04gh3HrlfNPwMtc&redir_esc=y#v=onepage&q&f=false} {\emph{Arabic culture: through its language and literature}}.
\newblock Taylor \& Francis.

\bibitem[{Baker(2018)}]{baker2018other}
Mona Baker. 2018.
\newblock \href {https://www.taylorfrancis.com/books/mono/10.4324/9781315619187/words-mona-baker} {\emph{In other words: A coursebook on translation}}.
\newblock Routledge.

\bibitem[{Belkebir and Habash(2021)}]{Belkebir2021AutomaticET}
Riadh Belkebir and Nizar Habash. 2021.
\newblock \href {https://api.semanticscholar.org/CorpusID:237532145} {Automatic error type annotation for arabic}.
\newblock In \emph{Conference on Computational Natural Language Learning}.

\bibitem[{Brosh(2013)}]{brosh2013proverbs}
Hezi Brosh. 2013.
\newblock \href {https://www.researchgate.net/profile/Hezi-Brosh/publication/265421984_Proverbs_in_the_Arabic_language_classroom/links/5864377e08aebf17d3974442/Proverbs-in-the-Arabic-language-classroom.pdf} {Proverbs in the arabic language classroom}.
\newblock \emph{International Journal of Humanities and Social Science}, 3(5):19--29.

\bibitem[{Chacoto(1994)}]{chacoto1994estudo}
Luc{\'\i}lia Chacoto. 1994.
\newblock \emph{Estudo e formaliza{\c{c}}{\~a}o das propriedades l{\'e}xico-sint{\'a}cticas das express{\~o}es fixas proverbiais}.

\bibitem[{Chakrabarty et~al.(2022)Chakrabarty, Saakyan, Ghosh, and Muresan}]{chakrabarty-etal-2022-flute}
Tuhin Chakrabarty, Arkadiy Saakyan, Debanjan Ghosh, and Smaranda Muresan. 2022.
\newblock \href {https://doi.org/10.18653/v1/2022.emnlp-main.481} {{FLUTE}: Figurative language understanding through textual explanations}.
\newblock In \emph{Proceedings of the 2022 Conference on Empirical Methods in Natural Language Processing}, pages 7139--7159, Abu Dhabi, United Arab Emirates. Association for Computational Linguistics.

\bibitem[{Chakrabarty et~al.(2023)Chakrabarty, Saakyan, Winn, Panagopoulou, Yang, Apidianaki, and Muresan}]{chakrabarty-etal-2023-spy}
Tuhin Chakrabarty, Arkadiy Saakyan, Olivia Winn, Artemis Panagopoulou, Yue Yang, Marianna Apidianaki, and Smaranda Muresan. 2023.
\newblock \href {https://doi.org/10.18653/v1/2023.findings-acl.465} {{I} spy a metaphor: Large language models and diffusion models co-create visual metaphors}.
\newblock In \emph{Findings of the Association for Computational Linguistics: ACL 2023}, pages 7370--7388, Toronto, Canada. Association for Computational Linguistics.

\bibitem[{Chen et~al.(2022)Chen, Chang, Zhang, Pu, Chen, Zhang, Xi, Chen, and Su}]{chen-etal-2022-probing}
Weijie Chen, Yongzhu Chang, Rongsheng Zhang, Jiashu Pu, Guandan Chen, Le~Zhang, Yadong Xi, Yijiang Chen, and Chang Su. 2022.
\newblock \href {https://doi.org/10.18653/v1/2022.acl-long.404} {Probing simile knowledge from pre-trained language models}.
\newblock In \emph{Proceedings of the 60th Annual Meeting of the Association for Computational Linguistics (Volume 1: Long Papers)}, pages 5875--5887, Dublin, Ireland. Association for Computational Linguistics.

\bibitem[{Chern et~al.(2024)Chern, Chern, Neubig, and Liu}]{chern2024can}
Steffi Chern, Ethan Chern, Graham Neubig, and Pengfei Liu. 2024.
\newblock \href {https://arxiv.org/abs/2401.16788} {Can large language models be trusted for evaluation? scalable meta-evaluation of llms as evaluators via agent debate}.
\newblock \emph{arXiv preprint arXiv:2401.16788}.

\bibitem[{Chung et~al.(2024)Chung, Hou, Longpre, Zoph, Tay, Fedus, Li, Wang, Dehghani, Brahma et~al.}]{chung2024scaling}
Hyung~Won Chung, Le~Hou, Shayne Longpre, Barret Zoph, Yi~Tay, William Fedus, Yunxuan Li, Xuezhi Wang, Mostafa Dehghani, Siddhartha Brahma, et~al. 2024.
\newblock \href {https://www.jmlr.org/papers/v25/23-0870.html} {Scaling instruction-finetuned language models}.
\newblock \emph{Journal of Machine Learning Research}, 25(70):1--53.

\bibitem[{Dankers et~al.(2022)Dankers, Lucas, and Titov}]{dankers-etal-2022-transformer}
Verna Dankers, Christopher Lucas, and Ivan Titov. 2022.
\newblock \href {https://doi.org/10.18653/v1/2022.acl-long.252} {Can transformer be too compositional? analysing idiom processing in neural machine translation}.
\newblock In \emph{Proceedings of the 60th Annual Meeting of the Association for Computational Linguistics (Volume 1: Long Papers)}, pages 3608--3626, Dublin, Ireland. Association for Computational Linguistics.

\bibitem[{Dubey et~al.(2024)Dubey, Jauhri, Pandey, Kadian, Al-Dahle, Letman, Mathur, Schelten, Yang, Fan et~al.}]{dubey2024llama}
Abhimanyu Dubey, Abhinav Jauhri, Abhinav Pandey, Abhishek Kadian, Ahmad Al-Dahle, Aiesha Letman, Akhil Mathur, Alan Schelten, Amy Yang, Angela Fan, et~al. 2024.
\newblock \href {https://arxiv.org/abs/2407.21783} {The llama 3 herd of models}.
\newblock \emph{arXiv preprint arXiv:2407.21783}.

\bibitem[{Durmus et~al.(2023)Durmus, Nguyen, Liao, Schiefer, Askell, Bakhtin, Chen, Hatfield-Dodds, Hernandez, Joseph et~al.}]{durmus2023towards}
Esin Durmus, Karina Nguyen, Thomas~I Liao, Nicholas Schiefer, Amanda Askell, Anton Bakhtin, Carol Chen, Zac Hatfield-Dodds, Danny Hernandez, Nicholas Joseph, et~al. 2023.
\newblock \href {https://arxiv.org/abs/2306.16388} {Towards measuring the representation of subjective global opinions in language models}.
\newblock \emph{arXiv preprint arXiv:2306.16388}.

\bibitem[{Dwivedi et~al.(2023)Dwivedi, Lavania, and Modi}]{dwivedi-etal-2023-eticor}
Ashutosh Dwivedi, Pradhyumna Lavania, and Ashutosh Modi. 2023.
\newblock \href {https://doi.org/10.18653/v1/2023.emnlp-main.428} {{E}ti{C}or: Corpus for analyzing {LLM}s for etiquettes}.
\newblock In \emph{Proceedings of the 2023 Conference on Empirical Methods in Natural Language Processing}, pages 6921--6931, Singapore. Association for Computational Linguistics.

\bibitem[{El-Rahman(2020)}]{el2020practicability}
Tayyara~A El-Rahman. 2020.
\newblock \href {https://journals.sagepub.com/doi/full/10.1177/1362168819895253?casa_token=ASCKtbkDIRUAAAAA%3AgAVpA1bmmqqwV6Q3L3Z1obNauMsGdLzBLkmP770ieqSZnkQ3ChuB4JtTXZGKBZDS_ssf-02IOg} {The practicability of proverbs in teaching arabic language and culture}.
\newblock \emph{Language Teaching Research}.

\bibitem[{Elmitwally and Alsayat(2020)}]{elmitwally2020classification}
NOUH~SABRI Elmitwally and Ahmed Alsayat. 2020.
\newblock \href {https://www.academia.edu/download/64733848/8_JATIT.pdf} {Classification and construction of arabic corpus: figurative and literal}.
\newblock \emph{Journal of Theoretical and Applied Information Technology}, 98(19).

\bibitem[{Farghal(2021)}]{farghal2021animal}
Mohammed Farghal. 2021.
\newblock \href {https://www.researchgate.net/publication/350836416_Animal_Proverbs_in_Jordanian_Popular_culture} {Animal proverbs in jordanian popular culture: A thematic and translational analysis}.
\newblock \emph{Journal of English Literature and Language}, 2(1):1--8.

\bibitem[{Ferguson(2003)}]{ferguson2003diglossia}
Charles~A Ferguson. 2003.
\newblock Diglossia.
\newblock In \emph{The bilingualism reader}, pages 71--86. Routledge.

\bibitem[{Fussell and Moss(2008)}]{Fussell}
Susan Fussell and Mallie Moss. 2008.
\newblock \href {https://www.taylorfrancis.com/chapters/edit/10.4324/9781315805917-8/figurative-language-emotional-communication-susan-fussell-mallie-moss} {Figurative language in emotional communication}.

\bibitem[{Gentner(1983)}]{gentner1983structure}
Dedre Gentner. 1983.
\newblock \href {https://www.sciencedirect.com/science/article/abs/pii/S0364021383800093} {Structure-mapping: A theoretical framework for analogy}.
\newblock \emph{Cognitive Science}, 7(2):155--170.

\bibitem[{Gibbs and Colston(2012)}]{gibbs2012interpreting}
Raymond~W Gibbs and Herbert~L Colston. 2012.
\newblock \href {https://books.google.ae/books?hl=en&lr=&id=7zIkM2RyyNYC&oi=fnd&pg=PP5&dq=Interpreting+figurative+meaning&ots=kofHaXFEWz&sig=Nwk57Jx41yeGRPdVPPyGne2hPI8&redir_esc=y} {\emph{Interpreting figurative meaning}}.
\newblock Cambridge University Press.

\bibitem[{Gonz{\'a}lez~Rey(2002)}]{gonzalez2002isabel}
M{\textordfeminine}~Gonz{\'a}lez~Rey. 2002.
\newblock Isabel, la phras{\'e}ologie du fran{\c{c}}ais.

\bibitem[{Gorjian(2024)}]{gorjian2024translating}
Bahman Gorjian. 2024.
\newblock \href {https://citeseerx.ist.psu.edu/document?repid=rep1&type=pdf&doi=6564a40a448db224a0bebeac95648caeaa96fdf7} {Translating english proverbs into persian: A case of comparative linguistics}.
\newblock \emph{Comparative Linguistics}.
\newblock Ph.D. dissertation.

\bibitem[{Gupta et~al.(2023)Gupta, Song, and Anumanchipalli}]{gupta2023investigating}
Akshat Gupta, Xiaoyang Song, and Gopala Anumanchipalli. 2023.
\newblock \href {https://arxiv.org/abs/2309.08163} {Investigating the applicability of self-assessment tests for personality measurement of large language models}.
\newblock \emph{arXiv preprint arXiv:2309.08163}.

\bibitem[{Hamdan et~al.(2023)Hamdan, Al-Madanat, and Hamdan}]{hamdan2023connotations}
Hady~J Hamdan, Hanan Al-Madanat, and Wael Hamdan. 2023.
\newblock \href {https://psycholing-journal.com/index.php/journal/article/view/1297} {Connotations of animal metaphors in the jordanian context}.
\newblock \emph{Psycholinguistics}, 33(1):132--166.

\bibitem[{He et~al.(2022)He, Cheng, Li, Xie, and Xiao}]{he-etal-2022-pre}
Qianyu He, Sijie Cheng, Zhixu Li, Rui Xie, and Yanghua Xiao. 2022.
\newblock \href {https://doi.org/10.18653/v1/2022.acl-long.543} {Can pre-trained language models interpret similes as smart as human?}
\newblock In \emph{Proceedings of the 60th Annual Meeting of the Association for Computational Linguistics (Volume 1: Long Papers)}, pages 7875--7887, Dublin, Ireland. Association for Computational Linguistics.

\bibitem[{Holes(2004)}]{holes2004modern}
Clive Holes. 2004.
\newblock \href {https://books.google.ae/books?hl=en&lr=&id=8E0Rr1xY4TQC&oi=fnd&pg=PR10&dq=Modern+Arabic:+Structures,+functions,+and+varieties&ots=X3uBhklee9&sig=d7WhORnCxXk6GVk5ByMHoN8yIHk&redir_esc=y#v=onepage&q=Modern%20Arabic%3A%20Structures%2C%20functions%2C%20and%20varieties&f=false} {\emph{Modern Arabic: Structures, functions, and varieties}}.
\newblock Georgetown University Press.

\bibitem[{Jang et~al.(2023)Jang, Yu, and Frassinelli}]{jang2023figurative}
Hyewon Jang, Qi~Yu, and Diego Frassinelli. 2023.
\newblock \href {https://aclanthology.org/2023.findings-acl.622/} {Figurative language processing: A linguistically informed feature analysis of the behavior of language models and humans}.
\newblock In \emph{Findings of the Association for Computational Linguistics: ACL 2023}, pages 9816--9832.

\bibitem[{Ji et~al.(2024)Ji, Li, Paul, Paavola, Lin, Chen, O'Brien, Luo, Sch{\"u}tze, Tiedemann et~al.}]{ji2024emma}
Shaoxiong Ji, Zihao Li, Indraneil Paul, Jaakko Paavola, Peiqin Lin, Pinzhen Chen, Dayy{\'a}n O'Brien, Hengyu Luo, Hinrich Sch{\"u}tze, J{\"o}rg Tiedemann, et~al. 2024.
\newblock \href {https://arxiv.org/abs/2409.17892} {Emma-500: Enhancing massively multilingual adaptation of large language models}.
\newblock \emph{arXiv preprint arXiv:2409.17892}.

\bibitem[{Johnson et~al.(2022)Johnson, Pistilli, Men{\'e}dez-Gonz{\'a}lez, Duran, Panai, Kalpokiene, and Bertulfo}]{johnson2022ghost}
Rebecca~L Johnson, Giada Pistilli, Natalia Men{\'e}dez-Gonz{\'a}lez, Leslye Denisse~Dias Duran, Enrico Panai, Julija Kalpokiene, and Donald~Jay Bertulfo. 2022.
\newblock \href {https://arxiv.org/abs/2203.07785} {The ghost in the machine has an american accent: value conflict in gpt-3}.
\newblock \emph{arXiv preprint arXiv:2203.07785}.

\bibitem[{Kabra et~al.(2023)Kabra, Liu, Khanuja, Aji, Winata, Cahyawijaya, Aremu, Ogayo, and Neubig}]{kabra-etal-2023-multi}
Anubha Kabra, Emmy Liu, Simran Khanuja, Alham~Fikri Aji, Genta Winata, Samuel Cahyawijaya, Anuoluwapo Aremu, Perez Ogayo, and Graham Neubig. 2023.
\newblock \href {https://doi.org/10.18653/v1/2023.findings-acl.525} {Multi-lingual and multi-cultural figurative language understanding}.
\newblock In \emph{Findings of the Association for Computational Linguistics: ACL 2023}, pages 8269--8284, Toronto, Canada. Association for Computational Linguistics.

\bibitem[{Karoui et~al.(2015)Karoui, Benamara, Moriceau, Aussenac-Gilles, and Belguith}]{karoui2015towards}
Jihen Karoui, Farah Benamara, V{\'e}ronique Moriceau, Nathalie Aussenac-Gilles, and Lamia~Hadrich Belguith. 2015.
\newblock \href {https://hal.science/hal-01686512/} {Towards a contextual pragmatic model to detect irony in tweets}.
\newblock In \emph{53rd Annual Meeting of the Association for Computational Linguistics (ACL 2015)}, volume~2, pages 644--650. ACL: Association for Computational Linguistics.

\bibitem[{Kova{\v{c}} et~al.(2023)Kova{\v{c}}, Sawayama, Portelas, Colas, Dominey, and Oudeyer}]{kovavc2023large}
Grgur Kova{\v{c}}, Masataka Sawayama, R{\'e}my Portelas, C{\'e}dric Colas, Peter~Ford Dominey, and Pierre-Yves Oudeyer. 2023.
\newblock \href {https://arxiv.org/abs/2307.07870} {Large language models as superpositions of cultural perspectives}.
\newblock \emph{arXiv preprint arXiv:2307.07870}.

\bibitem[{Krippendorff(2011)}]{krippendorff2011alpha}
Klaus Krippendorff. 2011.
\newblock \href {https://repository.upenn.edu/asc_papers/43/} {Computing krippendorff's alpha-reliability}.
\newblock Technical report, University of Pennsylvania ScholarlyCommons.

\bibitem[{Kuhareva(2008)}]{kuhareva2008arabic}
EV~Kuhareva. 2008.
\newblock \href {https://archive.org/details/arabicproverbsor00burc/page/n3/mode/2up} {Arabic proverbs and sayings}.
\newblock \emph{Dictionary with lexical and phraseological comments. Moscow: AST: Vosto. Zapad}.

\bibitem[{Liu et~al.(2023)Liu, Koto, Baldwin, and Gurevych}]{liu2023multilingual}
Chen~Cecilia Liu, Fajri Koto, Timothy Baldwin, and Iryna Gurevych. 2023.
\newblock \href {https://arxiv.org/abs/2309.08591} {Are multilingual llms culturally-diverse reasoners? an investigation into multicultural proverbs and sayings}.
\newblock \emph{arXiv preprint arXiv:2309.08591}.

\bibitem[{Mieder(2004)}]{mieder2004proverbs}
W.~Mieder. 2004.
\newblock \href {https://books.google.ae/books?id=rfWFu5C-tL0C} {\emph{Proverbs: A Handbook}}.
\newblock Greenwood Folklore Handbooks. Bloomsbury Academic.

\bibitem[{Mieder(2007)}]{mieder2007proverbs}
Wolfgang Mieder. 2007.
\newblock \href {https://www.degruyter.com/} {\emph{Proverbs as cultural units or items of folklore}}.
\newblock na.

\bibitem[{Mieder(2021)}]{mieder2021innovative}
Wolfgang Mieder. 2021.
\newblock From innovative anti-proverbs to modern proverbs.
\newblock \emph{The Discoursal Use of Phraseological Units}, page~61.

\bibitem[{Mpouli(2017)}]{mpouli2017annotating}
Suzanne Mpouli. 2017.
\newblock \href {https://aclanthology.org/W17-7403.pdf} {Annotating similes in literary texts}.
\newblock In \emph{Proceedings of the 13th Joint ISO-ACL Workshop on Interoperable Semantic Annotation (ISA-13)}.

\bibitem[{Niculae and Danescu-Niculescu-Mizil(2014)}]{niculae2014brighter}
Vlad Niculae and Cristian Danescu-Niculescu-Mizil. 2014.
\newblock \href {https://aclanthology.org/D14-1215.pdf} {Brighter than gold: Figurative language in user generated comparisons}.
\newblock In \emph{Proceedings of the 2014 conference on empirical methods in natural language processing (EMNLP)}, pages 2008--2018.

\bibitem[{Ouyang et~al.(2022)Ouyang, Wu, Jiang, Almeida, Wainwright, Mishkin, Zhang, Agarwal, Slama, Ray et~al.}]{ouyang2022training}
Long Ouyang, Jeffrey Wu, Xu~Jiang, Diogo Almeida, Carroll Wainwright, Pamela Mishkin, Chong Zhang, Sandhini Agarwal, Katarina Slama, Alex Ray, et~al. 2022.
\newblock \href {https://proceedings.neurips.cc/paper_files/paper/2022/file/b1efde53be364a73914f58805a001731-Paper-Conference.pdf} {Training language models to follow instructions with human feedback}.
\newblock \emph{Advances in neural information processing systems}, 35:27730--27744.

\bibitem[{Prystawski et~al.(2022)Prystawski, Thibodeau, Potts, and Goodman}]{prystawski2022psychologically}
Ben Prystawski, Paul Thibodeau, Christopher Potts, and Noah~D Goodman. 2022.
\newblock \href {https://arxiv.org/abs/2209.08141} {Psychologically-informed chain-of-thought prompts for metaphor understanding in large language models}.
\newblock \emph{arXiv preprint arXiv:2209.08141}.

\bibitem[{Qian et~al.(2024)Qian, Altam, Alqurishi, and Souissi}]{qian2024cameleval}
Zhaozhi Qian, Faroq Altam, Muhammad Saleh~Saeed Alqurishi, and Riad Souissi. 2024.
\newblock \href {https://arxiv.org/abs/2409.12623} {Cameleval: Advancing culturally aligned arabic language models and benchmarks}.
\newblock \emph{arXiv preprint arXiv:2409.12623}.

\bibitem[{Rafailov et~al.(2024)Rafailov, Sharma, Mitchell, Manning, Ermon, and Finn}]{rafailov2024direct}
Rafael Rafailov, Archit Sharma, Eric Mitchell, Christopher~D Manning, Stefano Ermon, and Chelsea Finn. 2024.
\newblock \href {https://proceedings.neurips.cc/paper_files/paper/2023/file/a85b405ed65c6477a4fe8302b5e06ce7-Paper-Conference.pdf} {Direct preference optimization: Your language model is secretly a reward model}.
\newblock \emph{Advances in Neural Information Processing Systems}, 36.

\bibitem[{Reid et~al.(2024)Reid, Savinov, Teplyashin, Lepikhin, Lillicrap, Alayrac, Soricut, Lazaridou, Firat, Schrittwieser et~al.}]{reid2024gemini}
Machel Reid, Nikolay Savinov, Denis Teplyashin, Dmitry Lepikhin, Timothy Lillicrap, Jean-baptiste Alayrac, Radu Soricut, Angeliki Lazaridou, Orhan Firat, Julian Schrittwieser, et~al. 2024.
\newblock \href {https://arxiv.org/abs/2403.05530} {Gemini 1.5: Unlocking multimodal understanding across millions of tokens of context}.
\newblock \emph{arXiv preprint arXiv:2403.05530}.

\bibitem[{Rinasovna~Fattakhova et~al.(2019)Rinasovna~Fattakhova, Albertovna~Gimadeeva, Rishatovna~Galimova, and Akzamovich~Shaihullin}]{rinasovna2019proverbs}
Aida Rinasovna~Fattakhova, Alfiya Albertovna~Gimadeeva, Emma Rishatovna~Galimova, and Timur Akzamovich~Shaihullin. 2019.
\newblock \href {https://rals.scu.ac.ir/article_15089_6a1d659d5889d49aa55a2fac3999437d.pdf} {Proverbs in arabic: Definition, classification, outlook reflection}.
\newblock \emph{Journal of Research in Applied Linguistics}, 10(Proceedings of the 6th International Conference on Applied Linguistics Issues (ALI 2019) July 19-20, 2019, Saint Petersburg, Russia):521--528.

\bibitem[{Saakyan et~al.(2024)Saakyan, Kulkarni, Chakrabarty, and Muresan}]{saakyan2024v}
Arkadiy Saakyan, Shreyas Kulkarni, Tuhin Chakrabarty, and Smaranda Muresan. 2024.
\newblock V-flute: Visual figurative language understanding with textual explanations.
\newblock \emph{arXiv preprint arXiv:2405.01474}.

\bibitem[{Sanh et~al.(2021)Sanh, Webson, Raffel, Bach, Sutawika, Alyafeai, Chaffin, Stiegler, Scao, Raja et~al.}]{sanh2021multitask}
Victor Sanh, Albert Webson, Colin Raffel, Stephen~H Bach, Lintang Sutawika, Zaid Alyafeai, Antoine Chaffin, Arnaud Stiegler, Teven~Le Scao, Arun Raja, et~al. 2021.
\newblock \href {https://arxiv.org/abs/2110.08207} {Multitask prompted training enables zero-shot task generalization}.
\newblock \emph{arXiv preprint arXiv:2110.08207}.

\bibitem[{Sellam et~al.(2020)Sellam, Das, and Parikh}]{sellam-etal-2020-bleurt}
Thibault Sellam, Dipanjan Das, and Ankur Parikh. 2020.
\newblock \href {https://doi.org/10.18653/v1/2020.acl-main.704} {{BLEURT}: Learning robust metrics for text generation}.
\newblock In \emph{Proceedings of the 58th Annual Meeting of the Association for Computational Linguistics}, pages 7881--7892, Online. Association for Computational Linguistics.

\bibitem[{Shutova(2011)}]{Shutova2011ComputationalAT}
Ekaterina Shutova. 2011.
\newblock \href {https://api.semanticscholar.org/CorpusID:14270142} {Computational approaches to figurative language}.

\bibitem[{Shwartz and Dagan(2019)}]{shwartz-dagan-2019-still}
Vered Shwartz and Ido Dagan. 2019.
\newblock \href {https://doi.org/10.1162/tacl_a_00277} {Still a pain in the neck: Evaluating text representations on lexical composition}.
\newblock \emph{Transactions of the Association for Computational Linguistics}, 7:403--419.

\bibitem[{Singh et~al.(2024)Singh, Patidar, and Vig}]{singh2024translating}
Pushpdeep Singh, Mayur Patidar, and Lovekesh Vig. 2024.
\newblock \href {https://arxiv.org/abs/2406.14504} {Translating across cultures: Llms for intralingual cultural adaptation}.
\newblock \emph{arXiv preprint arXiv:2406.14504}.

\bibitem[{Tan and Jiang(2021)}]{tan-jiang-2021-bert}
Minghuan Tan and Jing Jiang. 2021.
\newblock \href {https://aclanthology.org/2021.ranlp-1.156} {Does {BERT} understand idioms? a probing-based empirical study of {BERT} encodings of idioms}.
\newblock In \emph{Proceedings of the International Conference on Recent Advances in Natural Language Processing (RANLP 2021)}, pages 1397--1407, Held Online. INCOMA Ltd.

\bibitem[{Team et~al.(2024)Team, Riviere, Pathak, Sessa, Hardin, Bhupatiraju, Hussenot, Mesnard, Shahriari, Ram{\'e} et~al.}]{team2024gemma}
Gemma Team, Morgane Riviere, Shreya Pathak, Pier~Giuseppe Sessa, Cassidy Hardin, Surya Bhupatiraju, L{\'e}onard Hussenot, Thomas Mesnard, Bobak Shahriari, Alexandre Ram{\'e}, et~al. 2024.
\newblock \href {https://arxiv.org/abs/2408.00118} {Gemma 2: Improving open language models at a practical size}.
\newblock \emph{arXiv preprint arXiv:2408.00118}.

\bibitem[{Tursunov(2022)}]{meliboevich2022influence}
Faezdzhon~Meliboevich Tursunov. 2022.
\newblock \href {https://doi.org/10.30853/phil20220081} {The influence of the cultural aspect on the translation of proverbs and idioms (a case study of the tajiki/persian, russian and english languages)}.
\newblock \emph{Filologicheskie nauki. Voprosy teorii i praktiki}, 15(2):616--620.

\bibitem[{van~der Maaten and Hinton(2008)}]{article-tsne}
Laurens van~der Maaten and Geoffrey Hinton. 2008.
\newblock \href {https://www.jmlr.org/papers/volume9/vandermaaten08a/vandermaaten08a.pdf?fbcl} {Viualizing data using t-sne}.
\newblock \emph{Journal of Machine Learning Research}, 9:2579--2605.

\bibitem[{Versteegh(2014)}]{versteegh2014arabic}
Kees Versteegh. 2014.
\newblock \href {https://books.google.ae/books?hl=en&lr=&id=RiarBgAAQBAJ&oi=fnd&pg=PR3&dq=%40book%7Bversteegh+2014+arabic,&ots=9irdK5g0n0&sig=eb7FOxZV481g2SHbuynL34DTCN8&redir_esc=y#v=onepage&q=%40book%7Bversteegh%202014%20arabic%2C&f=false} {\emph{Arabic language}}.
\newblock Edinburgh University Press.

\bibitem[{Wang et~al.(2024)Wang, Yang, Huang, Yang, Majumder, and Wei}]{wang2024multilingual}
Liang Wang, Nan Yang, Xiaolong Huang, Linjun Yang, Rangan Majumder, and Furu Wei. 2024.
\newblock \href {https://arxiv.org/abs/2402.05672} {Multilingual e5 text embeddings: A technical report}.
\newblock \emph{arXiv preprint arXiv:2402.05672}.

\bibitem[{Webster(2000)}]{webster2000arabic}
Shelia Webster. 2000.
\newblock Arabic proverbs and related forms.
\newblock \emph{De Proverbio: An Electronic Journal of International Proverb Studies}, 6(2).

\bibitem[{Yosef et~al.(2023)Yosef, Bitton, and Shahaf}]{yosef-etal-2023-irfl}
Ron Yosef, Yonatan Bitton, and Dafna Shahaf. 2023.
\newblock \href {https://doi.org/10.18653/v1/2023.findings-emnlp.74} {{IRFL}: Image recognition of figurative language}.
\newblock In \emph{Findings of the Association for Computational Linguistics: EMNLP 2023}, pages 1044--1058, Singapore. Association for Computational Linguistics.

\bibitem[{Zeng et~al.(2020)Zeng, Song, Su, Xie, Song, and Luo}]{zeng2020neural}
Jiali Zeng, Linfeng Song, Jinsong Su, Jun Xie, Wei Song, and Jiebo Luo. 2020.
\newblock \href {https://ojs.aaai.org/index.php/AAAI/article/view/6496} {Neural simile recognition with cyclic multitask learning and local attention}.
\newblock In \emph{Proceedings of the AAAI Conference on Artificial Intelligence}, volume~34, pages 9515--9522.

\bibitem[{Zeng et~al.(2023)Zeng, Yu, Gao, Meng, Goyal, and Chen}]{zeng2023evaluating}
Zhiyuan Zeng, Jiatong Yu, Tianyu Gao, Yu~Meng, Tanya Goyal, and Danqi Chen. 2023.
\newblock \href {https://arxiv.org/abs/2310.07641} {Evaluating large language models at evaluating instruction following}.
\newblock \emph{arXiv preprint arXiv:2310.07641}.

\bibitem[{Zhang et~al.(2019)Zhang, Kishore, Wu, Weinberger, and Artzi}]{zhang2019bertscore}
Tianyi Zhang, Varsha Kishore, Felix Wu, Kilian~Q Weinberger, and Yoav Artzi. 2019.
\newblock \href {https://arxiv.org/abs/1904.09675} {Bertscore: Evaluating text generation with bert}.
\newblock \emph{arXiv preprint arXiv:1904.09675}.

\end{thebibliography}

\appendix
 \clearpage
 \appendixpage
 \addappheadtotoc
% \numberwithin{figure}{section}
% \numberwithin{table}{section}
% %%%%%%%%%%%%%%%%%%%%%%%%%%%%%%%%%%%%%%%%%%%%%%%%%%%
% % ############################################################

 We offer an addition structured as follows:

\begin{itemize}
\item Development of Figurative Language \ref{apendx:development_figlang}
\item Linguistic Background of Arabic \ref{apendx:linguistics_background}

\item Data Statistics \ref{apendx:data_statistics_full}
%\item Prompt Design \ref{apendx:prompt_design}.
\item Themes of Arabic Proverbs \ref{apendx:themes_Ar.proverbs}

\item Automatic Evaluation \ref{apendx:automatic_evalaution}
\item Human Evaluation Metrics \ref{apendx:human_evaluation_metrics}
\item Quality Assurance of the Data \ref{apendx:quality_assurance}
%\item Inter Annotator Agreement \ref{apendx:iaa}

\end{itemize}

\section{Development of Figurative Language.}
\label{apendx:development_figlang}
Prior work on figurative language understanding has covered a wide range of topics, including simile detection and generation~\cite{niculae2014brighter, mpouli2017annotating, zeng2020neural}. Although language models can often recognize non-literal language and may attend to it less, they still struggle to fully capture the implied meaning of such phrases~\cite{shwartz-dagan-2019-still}. More recent studies have shifted toward tasks focused on \textit{comprehending} figurative language~\cite{chakrabarty-etal-2022-flute, he-etal-2022-pre, prystawski2022psychologically, jang2023figurative}. Several studies have explored whether knowledge of figurative meaning is encoded in the learned representations by investigating how well these models capture non-literal meanings~\cite{tan-jiang-2021-bert, chen-etal-2022-probing, dankers-etal-2022-transformer, jang2023figurative}. Additionally, efforts in dataset development have focused on more diverse tasks aimed at comprehending figurative language, such as recognizing textual entailment (RTE)~\cite{chakrabarty-etal-2022-flute, kabra-etal-2023-multi}, multilingual understanding\cite{liu2023multilingual}, and tasks involving figurative language beyond text~\cite{ yosef-etal-2023-irfl, chakrabarty-etal-2023-spy,saakyan2024v}. However, the lack of comprehensive datasets focusing on figurative language, particularly Arabic proverbs, limits the ability to thoroughly evaluate LLMs’ cultural awareness and their true understanding of culturally embedded non-literal expressions. 

\section{Linguistic Background of Arabic}
\label{apendx:linguistics_background}
Arabic is spoken by approximately 450 million people worldwide and is the sole or joint official language in over twenty Middle Eastern and African nations, including Morocco, Algeria, Mauritania, Tunisia, Libya, Egypt, and Sudan. MSA is a descendant of Classical Arabic (CA), which was used in the 6th century and is the language of the Quran. While MSA has evolved from CA, it has been significantly simplified to suit contemporary literary, poetic, and official discourse. Although its syntactic structure remains unchanged, MSA continues to evolve in vocabulary and phraseology. This pan-Arab variety is widely used in written forms and media, including news broadcasts, political speeches, and public ceremonies.
Arabic is a highly sophisticated and intricate language, characterized by its rich morphological and grammatical features~\cite{holes2004modern}. The complexity of the Arabic language stems from its orthographic, morphological, semantic, and syntactic systems \cite{aransa2015statistical}. \\Moreover, the Arabic writing system uses diacritics, which are small marks placed above or below letters. These marks, called "vocalization," help with the correct pronunciation and meanings of words. They are important for clearing up many ambiguities in the text, ensuring accurate understanding and interpretation \cite{almanea2021automatic}. For example, the orthographic ambiguity of the diacritic in the word \AR{\small{كتب}} without diacritics can be ambiguous. However, when diacritics are added, it can take on different meanings: \AR{\small{كَتَبَ}} (kataba): With a fatha on each consonant is a verb that means "he wrote." \AR{\small{كُتُب }}(kutub): With a damma on the first and second consonants is a noun which means "books." \AR{\small{كُتِبَ }}(kutiba): With a damma  on the first consonant and a kasra  on the second consonant is a verb in the passive voice, and means "it was written." These diacritics provide clarity and precision and ensure that the intended meaning is conveyed accurately. The complexity and diversity of Arabic create significant challenges for NLP. This is due to the language's rich morphology, diverse dialects, and intricate script, which includes diacritics \cite{abdul-mageed-etal-2020-nadi, Belkebir2021AutomaticET}. These factors make tasks such as text processing, speech recognition, and machine translation more difficult compared to other languages. Understanding and addressing these challenges is crucial for improving NLP applications for Arabic speakers. Arabic language exhibits considerable linguistic diversity, which includes numerous dialects spread across a vast geographical area~\cite{versteegh2014arabic} and is recognized as a diglossic language as it exhibits two varieties (MSA and various dialects)  which are used interchangeably \cite{ferguson2003diglossia}.  Proverbs, as succinct and poignant expressions, reflect the values, wisdom, and social norms prevalent within various Arabic-speaking communities. By including proverbs from a wide range of Arabic dialects, it increases the complexity of the Arabic language. They serve as a tool for understanding the historical, social, and cultural contexts from which they originate.

\section{Data Statistics.}
\label{apendx:data_statistics_full}
Table~\ref{tab:data_summary} shows full statistics of our dataset.
Table~\ref{example_dataset_all} shows examples of all proposed tasks.

 \section{Themes of Arabic proverbs}
 \label{apendx:themes_Ar.proverbs}
 Arabic uniquely maintains a well-defined link between past and present, classical and modern, in structure, grammar, vocabulary, and cultural traditions~\cite{versteegh2014arabic}. This linguistic and cultural continuity is best represented in proverbs, which are deeply rooted in Arab culture and identity~\cite{el2020practicability} and convey popular wisdom across generations~\cite{rinasovna2019proverbs}. Understanding them requires grasping the underlying cultural values and social norms they reflect, as their rich semantic layers cannot be derived from mere literal interpretation~\cite{ahmed2005educational}.
 The animal names take over most of the Arabic proverbs \cite{al2021cultural},  For instance, \textit{lion} symbolizes strength, courage, royalty, and ferocity in Arab culture. It  has multiple figurative meanings that are widely used in the proverbs such as \AR{\small{هذا الشبل من ذاك الأسد}} ("this little lion from that lion") which highlights the similarity between a father and his son, usually in a positive light. The proverb is used to emphasize the inherited qualities or behaviors that are apparent between generations. Also,the lion often signifies a large portion, as in the proverb \AR{\small{حصة الأسد}} ("the lion's share"), indicating that a person has received the most significant portion.\\
 Another common theme in Arabic proverbs is the use of \textit{camels}, underscoring the Bedouin lifestyle's deep connection with this animal. Camels are integral to the daily lives of Bedouins and farmers, particularly in arid and semi-arid regions \cite{al2021cultural}. They are used for riding during peace and war, transporting burdens, and providing essential resources like milk and meat. 
 
\begin{figure*}[t]
   \centering
\includegraphics[width=\linewidth]{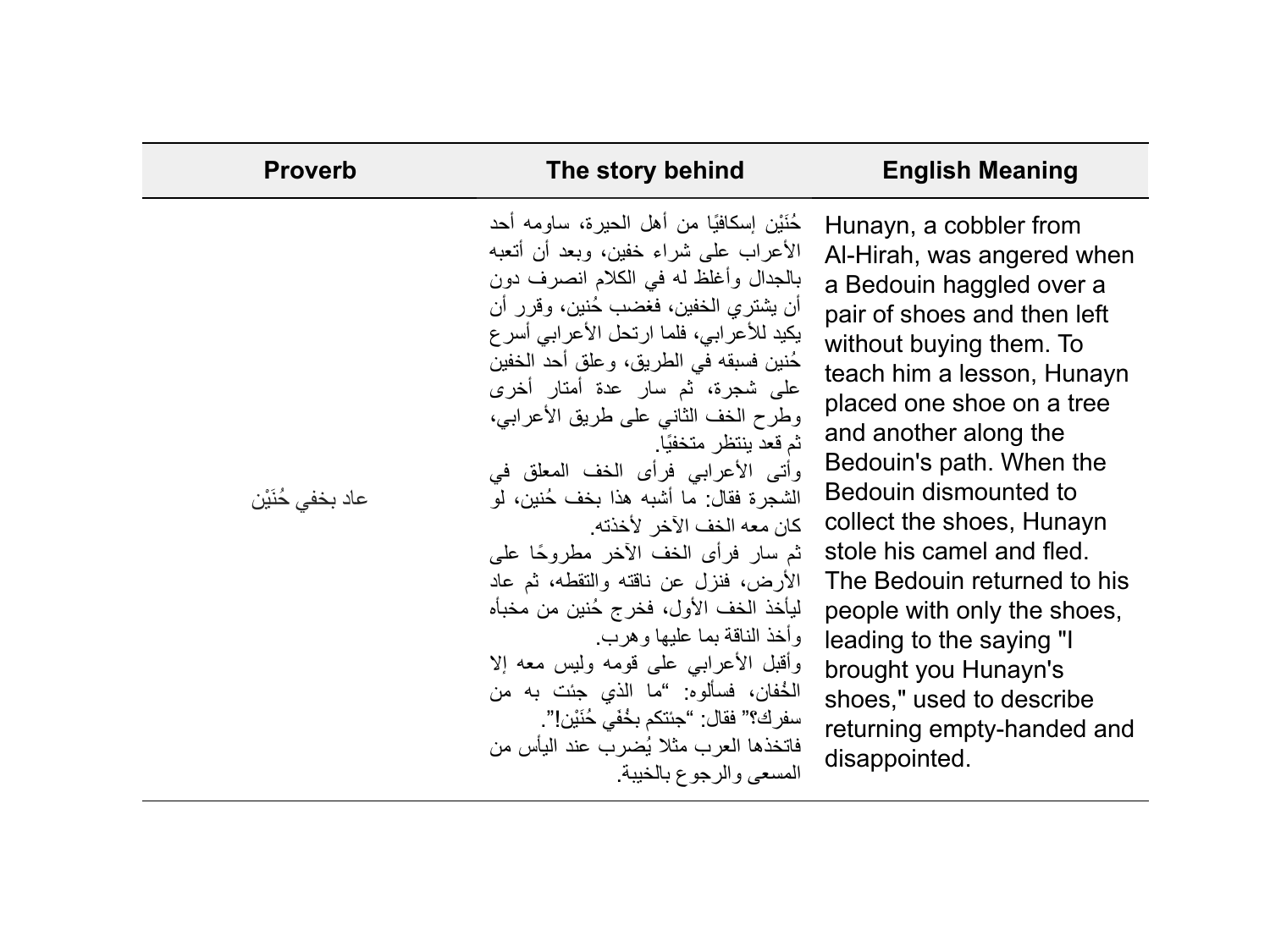}
   \caption{An MSA example illustrating the historical story behind a well known proverb.} \label{Figures/Story_Behind_Proverbs.pdf} 
 \end{figure*}

By comparing proverbs across the countries we investigate, we noticed that camel is a common theme and is used throughout proverbs with similar meanings, but with different words. For example, in Libya, they say \AR{\small{الجملْ ما يِحقّ عوجْ رقْبْتهْ}}, while in Levantine, they use \AR{\small{لو كان الجمل بيشُوفْ حردبنه كان بيقع بيفك رقبته}}. Both proverbs are used in describing a person who sees the flaws of others but does not see their own flaws. It is noticeable that there is a significant similarity in the proverbs across different Arab countries, whether in the East or the West, both in their wording and their meaning. This similarity stems from the strong cultural connections and the ongoing communication between these countries.\\
More examples of the camel theme are listed in table~\ref{camel-theme}.\\
 Some Arabic proverbs originate from true stories that occurred in the past. These proverbs have been passed down through generations and are still widely used today. These historical anecdotes provide context and meaning to the proverbs, highlighting the practical lessons and experiences of earlier times. For instance, the story behind a well-known MSA proverb \AR{\small{ على نفسها جَنَتْ براقش}}. This proverb dates back to the Pre-Islamic era. Baraqaš was a dog who inadvertently led enemy soldiers to her people through her barking. Thus, it is said, "Baraqaš brought it upon her people" or "She brought it upon herself." This ancient proverb is used to describe someone who harms themselves by their own actions; an illustrated example is provided in figure~\ref{Figures/Story_Behind_Proverbs.pdf}.\\

\section{Automatic Evaluation}
\label{apendx:automatic_evalaution}
Full results including Precision and Recall scores can be found in Table~\ref{Bert_score}.

  \section{Human Evaluation Metrics}
  \label{apendx:human_evaluation_metrics}
In our evaluation of LLMs on the \textit{Jawaher} dataset, we assess them on two main tasks: translation and providing explanations of proverbs across 20 Arabic varieties in both English and Arabic.\\
For the \textbf{translation task}, we evaluate the models' ability to provide the English equivalents of the proverbs. To measure the capabilities of these models to understand figurative language such as proverbs and to provide idiomatic translations, we assess the models on two main scales: \textit{accuracy} and \textit{idiomaticity}, each rated on a scale from 1 to 5. A summary of the metric in table~\ref{tab:eval_metic}\\
\textbf{Accuracy} ensures that the meaning is correct and how well the translation conveys the exact intent and cultural context with precise language, ensuring complete accuracy and clarity. \\
5: The translation fully captures the original proverb's meaning, intent, and cultural context with precise language, ensuring complete accuracy and clarity.\\
4: The translation is mostly accurate, capturing the original meaning and cultural context well, but it may miss some subtle nuances or minor details without significantly altering the overall meaning.\\
3: The translation conveys the general meaning of the proverb but lacks some nuance or cultural context, resulting in minor inaccuracies or partial misinterpretations.\\
2: The translation captures some elements of the original meaning, but it lacks important details or cultural context, leading to noticeable inaccuracies or a partial misinterpretation of the proverb.\\
1: The translation fails to convey the correct meaning or completely misrepresents the original proverb's intent and cultural context, or the translation is missing altogether.\\
The other criteria we assess is the idiomaticity of the translated proverbs. \\
\textbf{Idiomaticity} refers to how naturally the translation fits into the linguistic and cultural norms of the target language. A proverb should sound like something a native speaker would say, using phrases, metaphors, or expressions that are common in the target language. An idiomatic translation may diverge from a literal translation to better match the figurative or stylistic nature of the original proverb. Since proverbs are idiomatic expressions, their translations must feel natural in the target language. The translation should sound like a proverb in the target language rather than a forced, awkward sentence.\\
5: The translation is completely natural, idiomatic, and would be readily recognized as a proverb by native speakers.\\
4: The translation is almost natural and idiomatic, with only slight room for improvement.\\
3: The translation is generally understandable but contains notable awkwardness or unnatural phrasing.\\
2: The translation is somewhat understandable but remains somewhat awkward and far from idiomatic.\\
1: The translation is completely awkward or unnatural and is not idiomatic in the target language.\\
The second task in our evaluation is the \textbf{explanation} of the proverbs in both Arabic and English. We evaluated the models' outputs across four metrics: \textit{clarity}, \textit{correctness}, \textit{depth and detail}, and \textit{cultural relevance and sensitivity}.\\
\textbf{Clarity}: 
By \textit{clarity} we assess how clearly the meaning of the proverb is explained. The explanation should be understandable, and the evaluator should not need further clarification.\\
5: The explanation is perfectly clear and easy to understand, with no confusion.\\
4: The explanation is mostly clear and easy to follow, with only minor areas needing clarification.\\
3: The explanation is somewhat clear, but there are one or two areas that are confusing or lack detail.\\
2: The explanation is somewhat unclear, with several confusing or ambiguous points.\\
1: The explanation is very unclear or confusing, making it difficult to understand the meaning of the proverb.\\

\textbf{Correctness}: 
\textit{Correctness} criterion evaluates the accuracy of the explanation. It should reflect the actual meaning of the proverb without misinterpretation.\\
5: The explanation is completely correct and accurately conveys the meaning of the proverb.\\
4: The explanation is mostly correct, with only minor inaccuracies.\\
3: The explanation is partially correct but misses some key aspects or provides an incomplete interpretation.\\
2: The explanation is mostly incorrect, with significant inaccuracies or misinterpretations.\\
1: The explanation is incorrect, misinterpreting the proverb's meaning.\\

\textbf{Depth and Detail}
\textit{Depth and details} criterion measures how comprehensive the explanation is. It should cover any underlying meanings, cultural nuances, and possible interpretations, offering more than just a surface-level description.\\
5: The explanation is highly detailed, providing a thorough understanding of the proverb's meaning, cultural context, and potential interpretations.\\
4: The explanation is fairly detailed, covering cultural nuances or different interpretations with only minor omissions.\\
3: The explanation gives a basic understanding but does not delve deeply into cultural nuances or multiple interpretations.\\
2: The explanation provides some details but lacks sufficient depth or misses key elements.\\
1: The explanation is overly simplistic, lacking depth and failing to cover important aspects of the proverb.\\

 \textbf{Cultural Relevance and Sensitivity} 
 This criterion assesses whether the explanation acknowledges any cultural context necessary to fully understand the proverb. Proverb explanations should consider whether the audience understands the cultural or historical background if needed.\\
 5: The explanation fully considers cultural context, making it highly relevant and meaningful to the target audience.\\
 4: The explanation is culturally relevant, addressing most of the important cultural or contextual elements.\\
 3: The explanation somewhat considers the cultural context but could do more to connect with the audience’s cultural understanding.\\
 2: The explanation makes only minimal reference to cultural relevance, leaving important aspects unaddressed.\\
 1: The explanation does not account for the cultural context or gives an inappropriate explanation that does not resonate with the target audience.\\
 
Table~\ref{tab:eval_metic} outlines the full criteria of the evaluation metrics.

 \section{Quality Assurance of the Data}
 \label{apendx:quality_assurance}
 To ensure the high quality of our dataset and the subsequent evaluation process, we implemented a dual annotation procedure. Three native-speaking expert linguists meticulously verified the data by first reviewing the information gathered from online sources. They checked for language accuracy, clarity of explanations, and corrected any grammatical errors or other issues present in the data.
For the translation tasks, dual annotation was employed to assess the quality of translations comprehensively. This involved reviewing the English equivalents, meanings, and explanations to ensure consistency and accuracy across all tasks.
After completing these quality assurance steps, we proceeded with our evaluation.

\clearpage

%\usepackage{multirow}
%\usepackage{graphicx}

\begin{table}[]
\centering
\captionsetup{width=0.9\textwidth}
\renewcommand{\arraystretch}{1.5}
\resizebox{0.9\textwidth}{!}{%

\begin{tabular}{crll}
\hline
\textbf{Variety} &
  \multicolumn{1}{c}{\textbf{Proverb}} &
  \multicolumn{1}{c}{\textbf{English Translation}} &
  \multicolumn{1}{c}{\textbf{Shared Meaning}} \\ \hline
Algeria & \AR{\small{اللّي تلمّه النّملة في عام يأكله الجمل في لُقمة}} &
  \begin{tabular}[c]{@{}l@{}}What an ant collects in a year, a \\ \textit{camel} eats in one bite.\end{tabular} &
   \\
Egypt &
  \AR{\small{اللّي تجمعه النّملة في سنة يأخده الجمل في خُفّه}} &
  \begin{tabular}[c]{@{}l@{}}What an ant gathers in a year, a \\ \textit{camel} takes in its hoof.\end{tabular} & \multirow{6}{*}{\begin{tabular}[c]{@{}c@{}}These proverbs emphasize the idea that small, long\\-term efforts or savings can easily be taken or \\ consumed by something much larger in an instant.\end{tabular}} \\
Iraq &
  \AR{\small{اللّي تجمعه النّملة بسنة يشيله البعير بخُفّه}} &
  \begin{tabular}[c]{@{}l@{}}What an ant gathers in a year, a \\ \textit{camel} carries in its hoof.\end{tabular} &
   \\
Libya &
  \AR{\small{اللّي تحوّشه النّملة في عامْ يأخذها الجمل في خُفّه}} &
  \begin{tabular}[c]{@{}l@{}}What an ant saves in a year, a \\ \textit{camel} takes in its hoof.\end{tabular} &
   \\
Qatar & \AR{\small{
  يجمعه العصفور في سنة وياكله الجمل في لقمة}} &
  \begin{tabular}[c]{@{}l@{}}What a bird gathers in a year, a \\ \textit{camel} eats in one bite.\end{tabular} &
   \\
Tunisia & \AR{\small{
  اللّي تلمّو النّملة في عام يهزّو الجمل الجمل في فم}} &
  \begin{tabular}[c]{@{}l@{}}What an ant collects in a year, a \\ \textit{camel} sways in its mouth.\end{tabular} &
   \\ \hline
\end{tabular}%
}
\caption{Proverbs about \textit{camels} from six Arab countries illustrating a shared concept that long-term savings can easily be consumed or lost in an instant. }
\label{camel-theme}
\end{table}

\newcolumntype{L}{>{\centering\arraybackslash}m{3cm}}

\label{sec:appendix}

\newpage

\begin{table*}[h]
\centering

\begin{tabular}{cll}
\hline
\textbf{Task} &
  \textbf{Measure} &
  \textbf{Scale} \\ \hline
\multirow{10}{*}{\centering\rotatebox{90}{Translation}} &
  \multirow{5}{*}{Accuracy} &
  \textbf{5}: Fully captures meaning, intent, and cultural context precisely. \\
 &  &
  \textbf{4}: Mostly accurate; minor nuances missed without altering overall meaning. \\
 &  &
  \textbf{3}: Conveys general meaning but lacks some nuances or context. \\
 &  &
  \textbf{2}: Captures some elements but lacks important details or context. \\
 &  &
  \textbf{1}: Fails to convey correct meaning or misrepresents intent; may be missing. \\ \cline{2-3} 
 & \multirow{5}{*}{Idiomaticity} &
  \textbf{5}: Completely natural and idiomatic; recognized by native speakers. \\
 &  &
  \textbf{4}: Almost natural; slight improvement possible. \\
 &  &
  \textbf{3}: Understandable but contains awkward phrasing. \\
 &  &
  \textbf{2}: Somewhat understandable but remains awkward. \\
 &  &
  \textbf{1}: Completely awkward; not idiomatic. \\ \hline
\multirow{20}{*}{\centering\rotatebox{90}{Explanation}} &
  \multirow{5}{*}{Clarity} &
  \textbf{5}: Perfectly clear and easy to understand. \\
 &  &
  \textbf{4}: Mostly clear; minor clarification needed. \\
 &  &
  \textbf{3}: Somewhat clear; some confusing areas. \\
 &  &
  \textbf{2}: Unclear; several confusing points. \\
 &  &
  \textbf{1}: Very unclear; difficult to understand. \\ \cline{2-3} 
 & \multirow{5}{*}{Correctness} &
  \textbf{5}: Completely correct; accurately conveys meaning. \\
 &  &
  \textbf{4}: Mostly correct; minor inaccuracies. \\
 &  &
  \textbf{3}: Partially correct; misses key aspects. \\
 &  &
  \textbf{2}: Mostly incorrect; significant inaccuracies. \\
 &  &
  \textbf{1}: Incorrect; misinterprets meaning. \\ \cline{2-3} 
 & \multirow{5}{*}{Depth and Detail} &
  \textbf{5}: Highly detailed; thorough understanding. \\
 &  &
  \textbf{4}: Fairly detailed; minor omissions. \\
 &  &
  \textbf{3}: Basic understanding; lacks depth. \\
 &  &
  \textbf{2}: Lacks depth; misses key elements. \\
 &  &
  \textbf{1}: Overly simplistic; lacks important aspects. \\ \cline{2-3} 
 & \multirow{5}{*}{\parbox{2.8cm}{Cultural Relevance and Sensitivity}} &
  \textbf{5}: Fully considers cultural context; highly relevant. \\
 &  &
  \textbf{4}: Culturally relevant; addresses most elements. \\
 &  &
  \textbf{3}: Somewhat considers cultural context; could connect better. \\
 &  &
  \textbf{2}: Minimal cultural relevance; important aspects unaddressed. \\
 &  &
  \textbf{1}: Does not account for cultural context; inappropriate explanation. \\ \hline
\end{tabular}
\caption{A summary of the evaluation metrics for translation and explanation tasks of our work}
\label{tab:eval_metic}
\end{table*}

\begin{table*}[h]
\small % Reduce the font size
\begin{center}

\resizebox{0.8\textwidth}{!}{%
\begin{tabular}{m{1cm}m{2cm}m{3.5cm}m{3.5cm}m{3.5cm}}
 \hline
 \textbf{Variety} & \textbf{Example} & \textbf{En Equivalent} & \textbf{Ar. Explanation} & \textbf{En. Explanation} \\
 \hline

\textbf{ALG} &
\multicolumn{1}{m{2cm}}{ \AR{\small{ حوحو يشكر روحو }}} &
\multicolumn{1}{m{3.5cm}}{Don't brag about yourself let others praise you} &
\multicolumn{1}{m{3.5cm}}{ \AR{\small{يقال لمن يمدح نفسه ويشكرها تنمرا عليه حوحو يشكر روحو}}} &
\multicolumn{1}{m{3.5cm}}{Used to mock someone who praises themselves excessively.} \\\cline{1-5}

\textbf{BHR} &
\multicolumn{1}{m{2cm}}{ \AR{\small{إنفخ يا شريم قال ماكو برطم  }}} &
\multicolumn{1}{m{3.5cm}}{You can not escape bad luck} &
\multicolumn{1}{m{3.5cm}}{ \AR{\small{الشخص الذي لا يفهم ما تقول له، او تشرح له.}}} &
\multicolumn{1}{m{3.5cm}}{When someone is trying to explain something to a person who can't understand.} \\\cline{1-5}

\textbf{EGY} &
\multicolumn{1}{m{2cm}}{ \AR{\small{آخرة خدمة الغُز علقة}}} &
\multicolumn{1}{m{3.5cm}}{The end of favor is denial.} &
\multicolumn{1}{m{3.5cm}}{ \AR{\small{يقال للدلالة على نكران الجميل، ومقابلة الإحسان بالشر}}} &
\multicolumn{1}{m{3.5cm}}{Used to describe ingratitude, where good deeds are met with harm or betrayal.} \\\cline{1-5}

\textbf{IRQ} &
\multicolumn{1}{m{2cm}}{ \AR{\small{يتعلّم الحجامة بروس اليتامة}}} &
\multicolumn{1}{m{3.5cm}}{A burnt child dreads the fire.} &
\multicolumn{1}{m{3.5cm}}{ \AR{\small{الشخص الذي يستغل الاشخاص الفقراء، لكي يقوم بعدة تجارب فاشلة}}} &
\multicolumn{1}{m{3.5cm}}{Refers to someone experimenting recklessly at the expense of the weak or poor.} \\\cline{1-5}

\textbf{JOR} &
\multicolumn{1}{m{2cm}}{ \AR{\small{أتبدلت غزلانها بقرودها}}} &
\multicolumn{1}{m{3.5cm}}{A falling master makes a standing servant} &
\multicolumn{1}{m{3.5cm}}{ \AR{\small{يضرب عند تغير الأحوال وحلول الرديء مكان الجيد}}} &
\multicolumn{1}{m{3.5cm}}{Describes a situation where good circumstances are replaced by bad ones.} \\\cline{1-5}

\textbf{KWT} &
\multicolumn{1}{m{2cm}}{ \AR{\small{حاط دوبَه دُوبي}}} &
\multicolumn{1}{m{3.5cm}}{He picks on me} &
\multicolumn{1}{m{3.5cm}}{ \AR{\small{جعلني محط اهتمامه في الشجار والمجادلة}}} &
\multicolumn{1}{m{3.5cm}}{Describes someone who targets another person in arguments or disputes.} \\\cline{1-5}

\textbf{LBN} &
\multicolumn{1}{m{2cm}}{ \AR{\small{الي بكبِّر لقمتو، بِغِصّ فيا.}}} &
\multicolumn{1}{m{3.5cm}}{Don't live beyond your means} &
\multicolumn{1}{m{3.5cm}}{ \AR{\small{ يضرب في التّحذير من تخطّي الاعتدال.}}} &
\multicolumn{1}{m{3.5cm}}{This proverb urges moderation.} \\\cline{1-5}

\textbf{LIB} &
\multicolumn{1}{m{2cm}}{ \AR{\small{اضرب القطوس، تخاف العروس}}} &
\multicolumn{1}{m{3.5cm}}{beat the dog before the lion} &
\multicolumn{1}{m{3.5cm}}{ \AR{\small{أي اضرب الضعيف، ترهب القوي}}} &
\multicolumn{1}{m{3.5cm}}{Strike the weak, intimidate the strong.} \\\cline{1-5}

\textbf{MRT} &
\multicolumn{1}{m{2cm}}{ \AR{\small{ الدني ماجاهَ حواش}}} &
\multicolumn{1}{m{3.5cm}}{Do not spare anything in this life} &
\multicolumn{1}{m{3.5cm}}{ \AR{\small{من ينفق من ما اعطاه الله بيسر وبدونِ تقتير}}} &
\multicolumn{1}{m{3.5cm}}{Encourages living freely and without excessive restraint.} \\\cline{1-5}

\textbf{MOR} &
\multicolumn{1}{m{2cm}}{ \AR{\small{اش عرف الحمير في سكينجبير}}} &
\multicolumn{1}{m{3.5cm}}{Casting pearls before swine} &
\multicolumn{1}{m{3.5cm}}{ \AR{\small{جهل قيمة الاشياء الثمينة التي لا يقدرها حق قدره }}} &
\multicolumn{1}{m{3.5cm}}{When valuable things are wasted on those who cannot appreciate them.} \\\cline{1-5}

\textbf{MSA} &
\multicolumn{1}{m{2cm}}{ \AR{\small{تجري الرياح بما لا تشتهي السّفن}}} &
\multicolumn{1}{m{3.5cm}}{Things don’t always go as they’re planned.} &
\multicolumn{1}{m{3.5cm}}{ \AR{\small{أنَّ الحياة لا تسير دائما وفق رغبات الإنسان}}} &
\multicolumn{1}{m{3.5cm}}{Life doesn't always unfold according to one's desires or plans.} \\\cline{1-5}

\textbf{OMA} &
\multicolumn{1}{m{2cm}}{ \AR{\small{ برمة الشرك لا تثور}}} &
\multicolumn{1}{m{3.5cm}}{Two cooks spoil the cook} &
\multicolumn{1}{m{3.5cm}}{ \AR{\small{يقصد به المتزوجين الذين يريدون إدارة المنزل في نفس الوقت}}} &
\multicolumn{1}{m{3.5cm}}{Describes chaos that arises when too many people try to take control.} \\\cline{1-5}

\textbf{PAL} &
\multicolumn{1}{m{2cm}}{ \AR{\small{ أَعمَص وبِيِتجَعمَص}}} &
\multicolumn{1}{m{3.5cm}}{A beggar with a havana} &
\multicolumn{1}{m{3.5cm}}{ \AR{\small{ الشخص الذى يتعالى على الناس وهو ليس مثلهم}}} &
\multicolumn{1}{m{3.5cm}}{Someone who is acting superior but doesn't actually belong to that status.} \\\cline{1-5}

\textbf{QAT} &
\multicolumn{1}{m{2cm}}{ \AR{\small{ البيض اللى وقع على بيض فقشه}}} &
\multicolumn{1}{m{3.5cm}}{The rotten apple injures its neighbors} &
\multicolumn{1}{m{3.5cm}}{ \AR{\small{يقصد به الضعيف تظهر حقيقته اذا قابل ضعيفاً مثله}}} &
\multicolumn{1}{m{3.5cm}}{Weakness is exposed when confronted by another weak entity.} \\\cline{1-5}

\textbf{SAU} &
\multicolumn{1}{m{2cm}}{ \AR{\small{ما عنده إلا الخرطي}}} &
\multicolumn{1}{m{3.5cm}}{Be all talk and no action} &
\multicolumn{1}{m{3.5cm}}{ \AR{\small{الشخص الثرثار الذي لا يعمل ويقول ما لا يفعل }}} &
\multicolumn{1}{m{3.5cm}}{Someone who talks a lot but has no action behind their words.} \\\cline{1-5}

\textbf{SUD} &
\multicolumn{1}{m{2cm}}{ \AR{\small{أبيض جناح اسود مراح}}} &
\multicolumn{1}{m{3.5cm}}{Fine feathers do not make fine birds} &
\multicolumn{1}{m{3.5cm}}{ \AR{\small{يصف من يرتدي فاخر الثياب البيضاء، وتكون داره مسودة من القذارة}}} &
\multicolumn{1}{m{3.5cm}}{Describes someone who looks good on the outside but their true nature is the opposite.} \\\cline{1-5}

\textbf{SYR} &
\multicolumn{1}{m{2cm}}{ \AR{\small{ابن الديب ما بيتربّى}}} &
\multicolumn{1}{m{3.5cm}}{The apple doesn't fall far from the tree} &
\multicolumn{1}{m{3.5cm}}{ \AR{\small{يقال لمن يحاول أن يغيّر خصالاً أصيلة في إنسان ما}}} &
\multicolumn{1}{m{3.5cm}}{Describes how people often inherit their parent's traits, good or bad.} \\\cline{1-5}

\textbf{TUN} &
\multicolumn{1}{m{2cm}}{ \AR{\small{الجمل مايراش كربته يرا كربة غيرو}}} &
\multicolumn{1}{m{3.5cm}}{One does not see one’s own defects.} &
\multicolumn{1}{m{3.5cm}}{ \AR{\small{يقصد بها أن الشخص لا يرى عيوبه لكن يرى عيوب غيره}}} &
\multicolumn{1}{m{3.5cm}}{People who are quick to point out others' faults while ignoring their own.} \\\cline{1-5}

\textbf{UAE} &
\multicolumn{1}{m{2cm}}{ \AR{\small{اصبوعي في حلوجهم وصبوعهم في عيوني}}} &
\multicolumn{1}{m{3.5cm}}{Evil in return for good deed} &
\multicolumn{1}{m{3.5cm}}{ \AR{\small{عن نبذ الجحود، من يفعل الخير الكثير و لا يلقى سوى النكران}}} &
\multicolumn{1}{m{3.5cm}}{When good deeds are repaid with ungratefulness or evil actions.} \\\cline{1-5}

\textbf{YEM} &
\multicolumn{1}{m{2cm}}{ \AR{\small{تزين بالخلاخل، والبلا من داخل}}} &
\multicolumn{1}{m{3.5cm}}{Beauty is only skin deep} &
\multicolumn{1}{m{3.5cm}}{ \AR{\small{الرجل الذي يختار أن يتزوج امرأة جميلة، ولكنه لا ينظر إلى أخلاقها}}} &
\multicolumn{1}{m{3.5cm}}{Refers to a man who decides to marry a beautiful woman, but he does not care about her morals.} \\\cline{1-5}

\end{tabular}%
}
\end{center}
\caption{Examples from 20 Arabic varieties arranged in alphabetical order include the variety, Arabic proverbs, English equivalents, Arabic explanations, and English explanations}
\label{example_dataset_all}
\end{table*}

\begin{table*}[t!]
\parbox{.45\textwidth}{
\centering
\resizebox{\columnwidth}{!}{%
\begin{tabular}{lccccc}
\toprule
\textbf{Variety} & \textbf{No. Proverbs} & \textbf{Ar Explanation} & \textbf{En Explanation} & \textbf{En Equivalent}  \\
\midrule
\textbf{ALG} & $312$ & $153$ & $108$ & $54$  \\
\textbf{BAH} & $134$ & $112$ & $103$ & $25$  \\
\textbf{EGY} & $1,018$ & $781$ & $224$ & $314$ \\
\textbf{IRQ} & $304$ & $104$ & $104$ & $26$ \\
\textbf{JOR} & $398$ & $132$ & $127$ & $26$  \\
\textbf{KUW} & $126$ & $102$ & $102$ & $44$ \\
\textbf{LEB} & $495$ & $257$ & $111$ & $24$ \\
\textbf{LIB} & $390$ & $114$ & $109$ & $40$  \\
\textbf{MAU} & $360$ & $127$ & $112$ & $32$  \\
\textbf{MOR} & $507$ & $255$ & $159$ & $77$  \\
\textbf{MSA} & $604$ & $228$ & $199$ & $185$  \\
\textbf{OMA} & $182$ & $100$ & $100$ & $30$  \\
\textbf{PAL} & $1,078$ & $210$ & $121$ & $60$ \\
\textbf{QAT} & $161$ & $152$ & $139$ & $33$ \\
\textbf{SAU} & $302$ & $197$ & $111$ & $51$ \\
\textbf{SUD} & $228$ & $111$ & $102$ & $46$  \\
\textbf{SYR} & $1,723$ & $199$ & $102$ & $45$ \\
\textbf{TUN} & $1,221$ & $168$ & $126$ & $57$ \\
\textbf{UAE} & $212$ & $137$ & $135$ & $71$ \\
\textbf{YEM} & $282$ & $125$ & $106$ & $49$ \\
\midrule
\rowcolor{blue!5} \textbf{Total} & \textbf{$10,037$} & \textbf{$3,764$} & \textbf{$2,500$} & \textbf{$1,289$}  \\
\bottomrule
\end{tabular}%
}
\caption{Summary of data statistics for 20 Arabic varieties, listed alphabetically. The table includes the total \textsuperscript{$\star$}\texttt{No.}: Number of Arabic proverbs, their \textsuperscript{$\star$}\texttt{Ar}: Arabic explanations, \textsuperscript{$\star$}\texttt{En}: English explanations, and \textsuperscript{$\star$}\texttt{En}: English equivalents. }
\label{tab:data_summary}
}
\hfill
\parbox{.45\textwidth}{
\centering
\resizebox{\columnwidth}{!}{%
\begin{tabular}{llccc|ccc|ccc}
\toprule
\multirow{2}{*}{\textbf{Source}} &\multirow{2}{*}{\textbf{Models}} &\multicolumn{3}{c}{\textbf{English Equivalent}} &\multicolumn{3}{c}{\textbf{English Explanation}} &\multicolumn{3}{c}{\textbf{Arabic Explanation}}\\\cmidrule{3-5}\cmidrule{6-8}\cmidrule{9-11}
& &\textbf{P} &\textbf{R} &\textbf{F\textsubscript{1.0}} &\textbf{P} &\textbf{R} &\textbf{F\textsubscript{1.0}} &\textbf{P} &\textbf{R} &\textbf{F\textsubscript{1.0}} \\\toprule
\multirow{3}{*}{\textbf{Open}} 
&Llama $3.1$ & $87.28$~ & $91.67$~ & $89.42$~ &$86.83$~ &$91.68$~ &$89.19$~ &$67.31$~ &$66.94$~ &$67.05$~  \\
&Llama-$3.2$ &$87.41$~ &$91.66$~ &$89.48$~ &$86.92$~ &$91.66$~ &$89.23$~ &$65.88$~ &$66.50$~ &$66.10$~ \\
&Gemma$9$B-instruct &$88.81$~ &$92.76$~ &$90.74$~ &$88.53$~ &$92.85$~ &$90.64$~ &$67.23$~ &$67.35$~ &$67.22$~  \\\midrule
\multirow{4}{*}{\textbf{Closed}} 
&GPT-4o &$89.59$~ &$92.91$~ &$91.22$~ &$92.42$~ &$95.98$~ &$94.17$~ &$67.33$~ &$71.29$~ &$69.21$~ \\
&Gemini $1.5$ Pro
&$87.89$~ &$91.66$~ &$89.73$~ &$87.59$~ &$91.91$~ &$89.69$~ &$66.75$~ &$70.69$~ &$68.62$~ \\
&Cohere Command R+
&$88.05$~ &$92.24$~ &$90.09$~ &$87.65$~ &$92.33$~ &$89.93$~ &$65.97$~ &$70.99$~ &$68.34$~ \\
&Claude $3.5$  Sonnet& $90.53$~ & $90.76$~ & $90.64$~ &$81.34$~ &$86.07$~ &$83.62$~ &$67.27$~ &$71.04$~ &$69.06$~  \\
\bottomrule
\end{tabular}%
}
\caption{BERTScore table. T1: English equivalent, T2: English explanation, T3: Arabic explanation}
\label{Bert_score}
}
\end{table*}

\end{document}